\documentclass[10pt, a4paper, twocolumn]{article} % 10pt font size (11 and 12 also possible), A4 paper (letterpaper for US letter) and two column layout (remove for one column)

\usepackage[english]{babel} % English language hyphenation

\usepackage{microtype} % Better typography

\usepackage{amsmath,amsfonts,amsthm} % Math packages for equations

\usepackage{mathtools}

\usepackage[svgnames]{xcolor} % Enabling colors by their 'svgnames'

\usepackage{ragged2e}

\usepackage[small, labelfont=bf, up]{caption} % Custom captions under/above tables and figures

\usepackage{subcaption}

\usepackage{booktabs} % Horizontal rules in tables

\usepackage{lastpage} % Used to determine the number of pages in the document (for "Page X of Total")

\usepackage{graphicx} % Required for adding images

\usepackage{enumitem} % Required for customising lists
\setlist{noitemsep} % Remove spacing between bullet/numbered list elements

\usepackage{sectsty} % Enables custom section titles
\allsectionsfont{\usefont{OT1}{phv}{b}{n}} % Change the font of all section commands (Helvetica)

\usepackage{siunitx}

\usepackage{orcidlink}

\hypersetup{
    colorlinks,
    linkcolor={black},
    citecolor={MediumBlue},
    urlcolor={MediumBlue}
}

\usepackage{siunitx}

\usepackage{geometry} % Required for adjusting page dimensions

\usepackage[T1]{fontenc} % Output font encoding for international characters
\usepackage[utf8]{inputenc} % Required for inputting international characters

\usepackage{XCharter} % Use the XCharter font

\usepackage{fancyhdr} % Needed to define custom headers/footers
\fancypagestyle{firstpage}{ % Page style for the first page with the title
	\fancyhf{}
}

\newcommand{\authorstyle}[1]{{\large\usefont{OT1}{phv}{b}{n}\color{MidnightBlue}#1}} % Authors style (Helvetica)

\newcommand{\institution}[1]{{\footnotesize\usefont{OT1}{phv}{m}{sl}\color{Black}#1}} % Institutions style (Helvetica)

\usepackage{titling} % Allows custom title configuration

\newcommand{\HorRule}{\color{cyan}\rule{\linewidth}{1pt}} % Defines the gold horizontal rule around the title

\pretitle{
	\vspace{-50pt} % Move the entire title section up
	\HorRule\vspace{-20pt} % Horizontal rule before the title
	\fontsize{23}{26}\usefont{OT1}{phv}{b}{n}\selectfont % Helvetica
	\color{MidnightBlue} % Text colour for the title and author(s)
}

\posttitle{\par\vskip 10pt} % Whitespace under the title

\preauthor{} % Anything that will appear before \author is printed

\postauthor{ % Anything that will appear after \author is printed
	\vspace{0pt} % Space before the rule
	\par\HorRule % Horizontal rule after the title
	\vspace{-32pt} % Space after the title section
}

\usepackage{multicol}
\usepackage{lettrine} % Package to accentuate the first letter of the text (lettrine)
\usepackage{fix-cm}	% Fixes the height of the lettrine

\newcommand{\initial}[1]{ % Defines the command and style for the lettrine
	\lettrine[lines=3,findent=4pt,nindent=0pt]{% Lettrine takes up 3 lines, the text to the right of it is indented 4pt and further indenting of lines 2+ is stopped
		\color{cyan}% Lettrine colour
		{#1}% The letter
	}{}%
}

\usepackage{xstring} % Required for string manipulation

\newcommand{\lettrineabstract}[1]{
	\StrLeft{#1}{1}[\firstletter] % Capture the first letter of the abstract for the lettrine
	\initial{\firstletter}\StrGobbleLeft{#1}{1} % Print the abstract with the first letter as a lettrine
}

\usepackage[backend=bibtex,style=ieee,natbib=true]{biblatex} % Use the bibtex backend with the authoryear citation style (which resembles APA)

\usepackage[autostyle=true]{csquotes} % Required to generate language-dependent quotes in the bibliography
\title{\begin{flushleft}Augmenting cobots for sheet-metal SMEs with 3D object recognition and localisation\end{flushleft}}

\author{
	\authorstyle{Martijn Cramer~\orcidlink{0000-0002-1828-916X}, Yanming Wu~\orcidlink{0000-0002-8705-0902}, David De Schepper~\orcidlink{0000-0003-2090-0076} and Eric Demeester~\orcidlink{0000-0001-6866-3802}} % Authors
	\newline\newline % Space before institutions
	\institution{KU Leuven, Diepenbeek Campus, Dept. of Mechanical Engineering, Research unit ACRO, B-3000 Leuven, Belgium}\\
	\institution{Flanders Make @ KU Leuven, B-3001 Heverlee, Belgium}\\
}

\date{} % Add a date here if you would like one to appear underneath the title block, use \today for the current date, leave empty for no date

\begin{document}

% \maketitle % Print the title

\thispagestyle{firstpage} % Apply the page style for the first page (no headers and footers)

%-------------------------------------------------
%	ABSTRACT
%------------------------------------------------

\maketitle
\thispagestyle{empty}

\lettrineabstract{Sheet-metalworking small and medium-sized enterprises (SMEs) are increasingly challenged by high-mix--low-volume production, requiring them to handle small series and fluctuating orders. As the manufacturing landscape evolves, these businesses face mounting pressure to accommodate smaller production runs, faster delivery times, broader product variety, and unpredictable demand. This already challenging situation is further exacerbated by a shortage of skilled labour, making it even more difficult.}

Standard automation solutions tend to fall short for small-batch production, proving too rigid and costly. As a result, SMEs keep resorting to repetitive manual labour, driving up production costs and preventing tech-skilled workers from being employed to their full potential. While flexible cobotic solutions offer a promising alternative, adoption remains limited due to technological complexity, a lack of expertise in integration, and unclear return on investment (ROI) timelines.

The COOCK+ ROBUST project seeks to reduce those barriers by transforming cobots into mobile and reconfigurable production assistants, leveraging existing technologies. More specifically, participating metalworking companies will be supported in implementing (semi-)automated cobot-based solutions through a selection tool, several hands-on demonstrators, and the dissemination of knowledge via events, workshops and reports. This initiative is jointly spearheaded by the ACRO research unit of KU Leuven and sector organisation Sirris with the financial support of VLAIO.

\begin{figure}[htb]
	\centering
	\includegraphics[width=\linewidth]{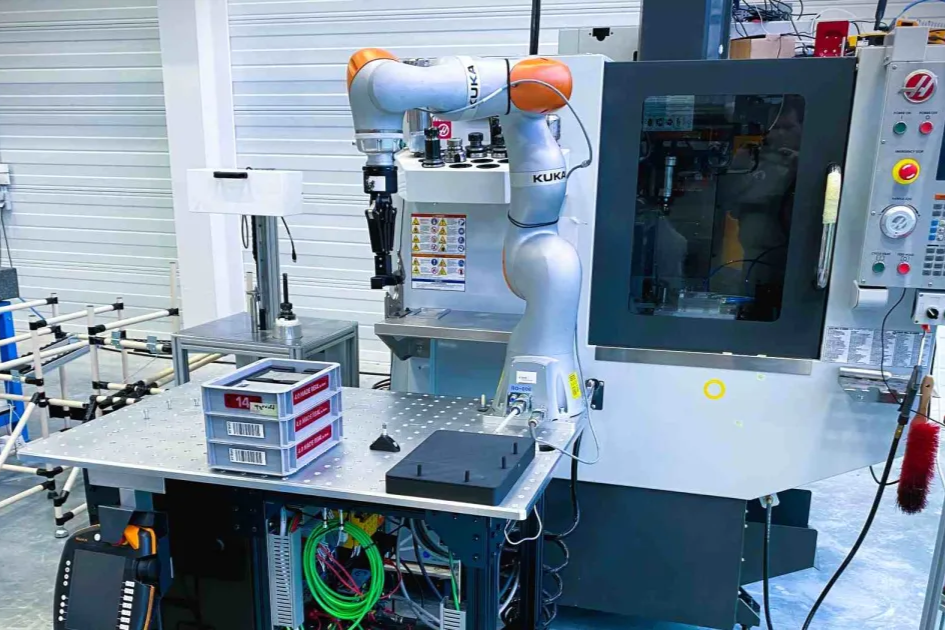}
	\caption{Mobile cobotic production assistant developed by Sirris, ready to (un)load a CNC machine~\cite{Sirris:2025}.}\label{fig:sirris_cobotic_production_assistant}
\end{figure}

In order for any cobotic production assistant to interact with the dynamic and unpredictable environment of a metal workshop, it must possess the fundamental ability of \emph{perception}. That is, the collaborative robot must be capable of identifying the workpieces, tools and machines in its vicinity, as well as determining their positions and orientations relative to a reference frame: processes known as \emph{object recognition} and \emph{localisation}. Ultimately, this will allow the (semi-)autonomous agent to engage in actions such as pick-and-place operations, machine tending, and other metalworking tasks.

This article addresses the potential and challenges of implementing these technologies in an industrial setting and outlines the key stages of the process---from a cobot scanning its environment with a depth camera and identifying relevant objects to translating these data into robot commands for tasks such as grasping. Insights gained from a past project, commissioned by one of ACRO's industrial partners, will serve as a practical use case and reference throughout. The software functions and algorithms discussed here are drawn from a commercial machine vision toolbox, extensively tested and validated across various industrial applications.

%------------------------------------------------
%	ARTICLE CONTENTS
%------------------------------------------------

\section{Computer vision terminology}\label{sec:terminology}

\emph{Object localisation} is concerned with estimating the pose of an object from sensor data, such as images or point clouds, relative to the camera's coordinate frame. It involves pinpointing the position and orientation of the object of interest in space. A method that determines an object's position using only X and Y coordinates, possibly including its orientation around the Z-axis, is referred to as \emph{2D pose estimation}. When the pose is instead defined using all six degrees of freedom---incorporating three positional and three rotational components---the algorithm performs \emph{3D pose estimation}. Throughout this article, object localisation and pose estimation will be used interchangeably.

\begin{figure}[htb]
	\centering
	\includegraphics[width=\linewidth]{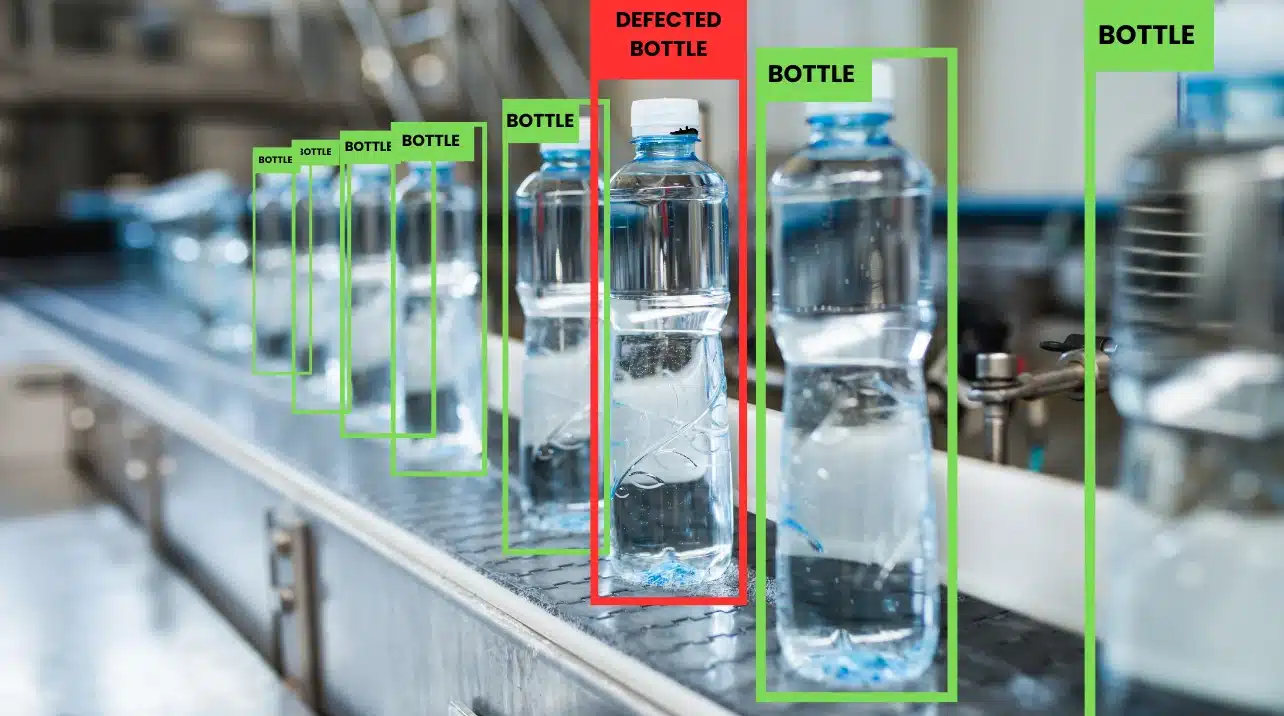}
	\caption{2D object detection of defective bottles using YOLO~\cite{EasyODM:2024}.}\label{fig:yolo_object_detection}
\end{figure}

An example of a 2D \emph{object detection}\footnote{It is important to note that object detection encompasses two key concepts. In addition to object localisation, it also includes \emph{object recognition}} algorithm, highly valued in academia for both its speed and accuracy, is YOLO~\cite{Wang:2022}. This deep-learning model is capable of analysing colour images or video streams, detecting all objects it has been trained to recognise, categorising them into their respective classes, and producing bounding boxes that encapsulate clusters of pixels belonging to the same object class, which involves classifying the detected object based on a predefined set of categories e.g., the different product types or states (raw material, processed, finished, rejected, packaged). These bounding boxes delineate the detected objects, hence providing information about their 2D poses with respect to the image's coordinate frame. Figure~\ref{fig:yolo_object_detection} illustrates an industrial application of YOLO in quality control, where it is deployed to detect and eliminate defective products from a bottle conveyor.
 
\begin{figure}[htb]
	\centering
	\includegraphics[width=\linewidth]{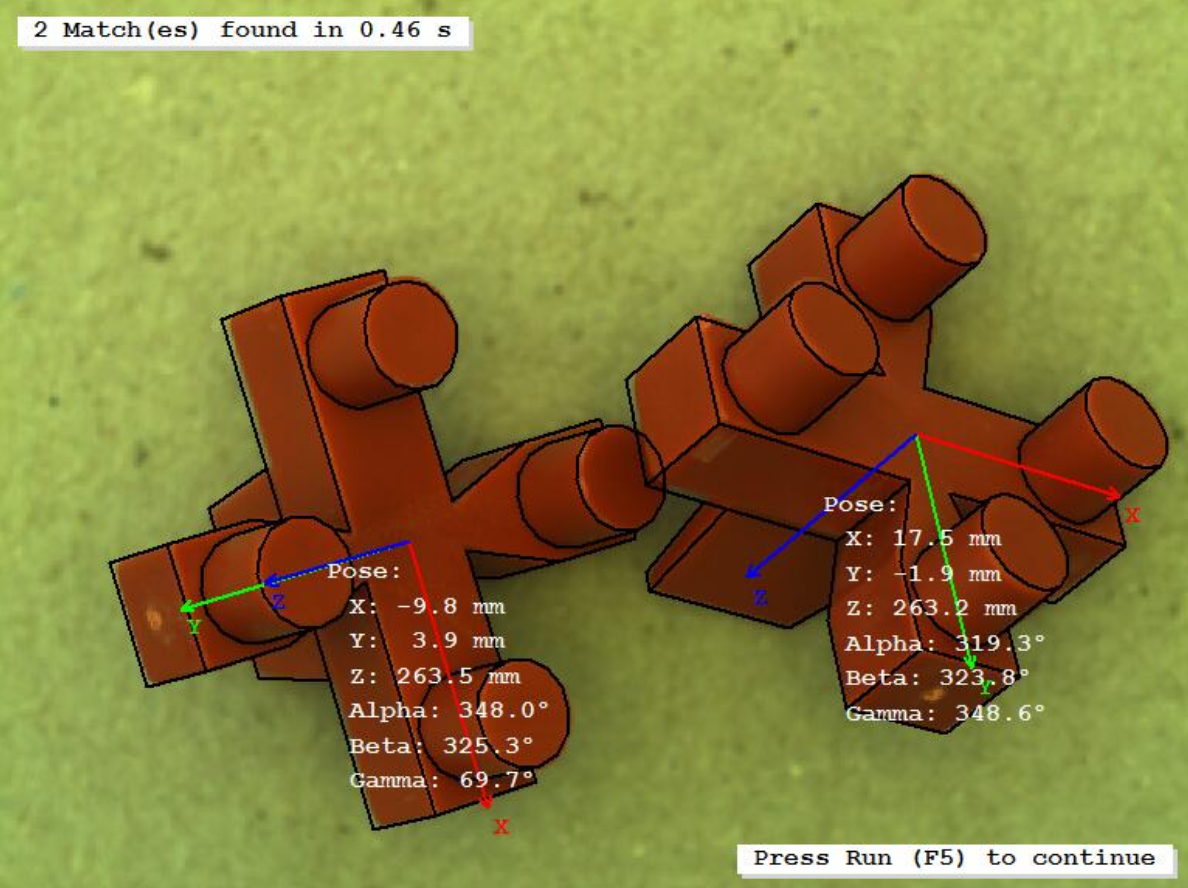}
	\caption{3D pose estimation using HALCON's shape-based matching algorithm~\cite[Ch. 4]{MVTec:2025a}.}\label{fig:halcon_shape_based_matching}
\end{figure}

For industrial applications requiring 3D object localisation, the commercial machine vision software MVTec HALCON~\cite{MVTec:2025} often is the go-to. It comprises a comprehensive library of standard computer vision algorithms featuring an integrated development environment (IDE). The following \emph{3D matching algorithms} in HALCON will be the focus of this article:
\begin{itemize}[leftmargin=*]
	\item \emph{3D primitives fitting} can be used to match primitive shapes e.g., planes, cylinders, etc. in 3D point clouds of the scene,

	\item \emph{3D surface-based matching}\footnote{In academic literature, surface-based matching is also known as point pair feature matching (PPFM)~\cite{Drost:2010}} can be used to match 3D CAD models of the workpieces in 3D point clouds of the scene,

	\item \emph{3D edge-supported surface-based matching} can be used to match 3D CAD models and edge information of the workpieces in 3D point clouds of the scene,

	\item \emph{3D shape-based matching} can be used to match 3D CAD models of the workpieces in 2D images of the scene (see Figure~\ref{fig:halcon_shape_based_matching}).
\end{itemize}
These techniques will be implemented and evaluated in the context of robotic handling tasks involving sheet-metal products, with the aim of identifying the most effective approach (or combination of approaches) for accurate and robust 3D object localisation and recognition in industrial environments.

%------------------------------------------------

\section{Practical relevance for sheet-metalworking SMEs}

The ability to recognise surrounding objects and determine their 6D poses could enhance a cobotic production assistant's flexibility and robustness in multiple ways.

\subsection{Less-constrained supply of parts}

The cobot could use its vision sensors to recognise and localise the parts from an unstructured (e.g., a disordered bin of workpieces) or semi-structured supply (e.g., a stack of semi-finished products) given that the input stream's location is roughly known relative to the cobot. As a result, workpieces no longer need to be presented to the cobot in predetermined positions and orientations, easing the constraints on how parts are supplied.

\begin{figure}[htb]
     \centering
     \begin{subfigure}[b]{0.47\textwidth}
         \includegraphics[width=\textwidth]{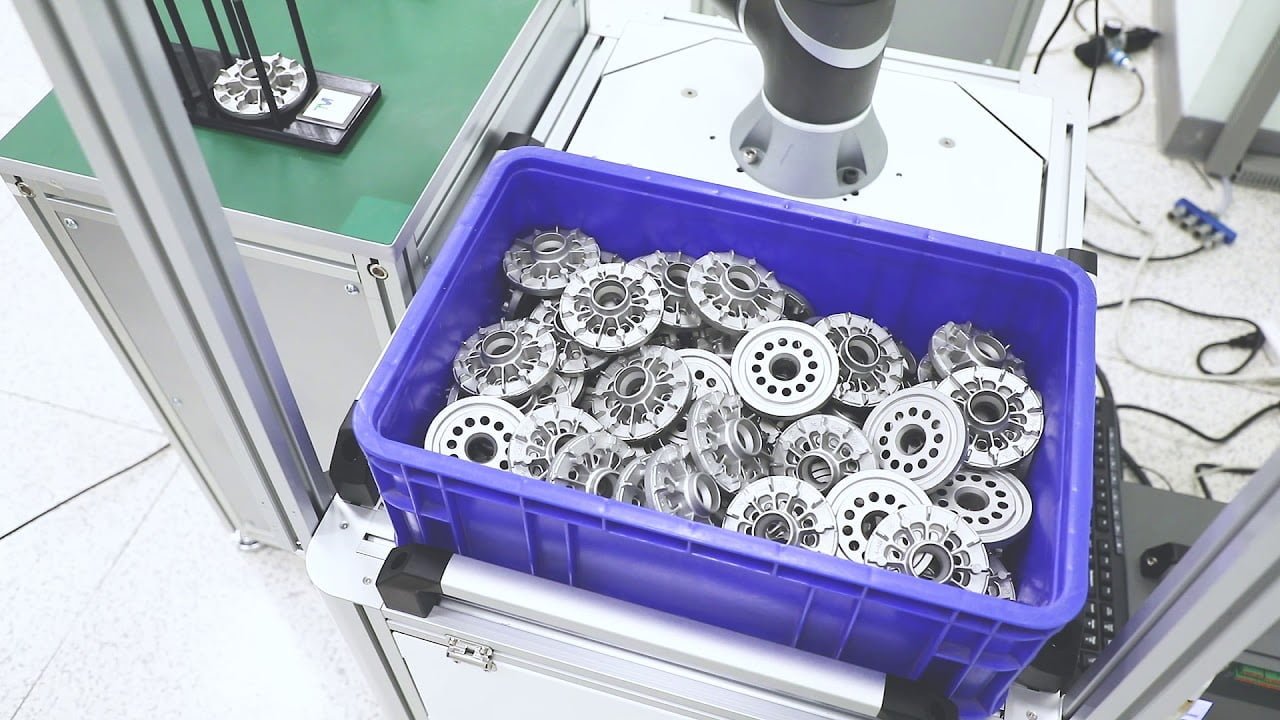}
     \end{subfigure}
	 \par\smallskip
     \begin{subfigure}[b]{0.47\textwidth}
         \includegraphics[width=\textwidth]{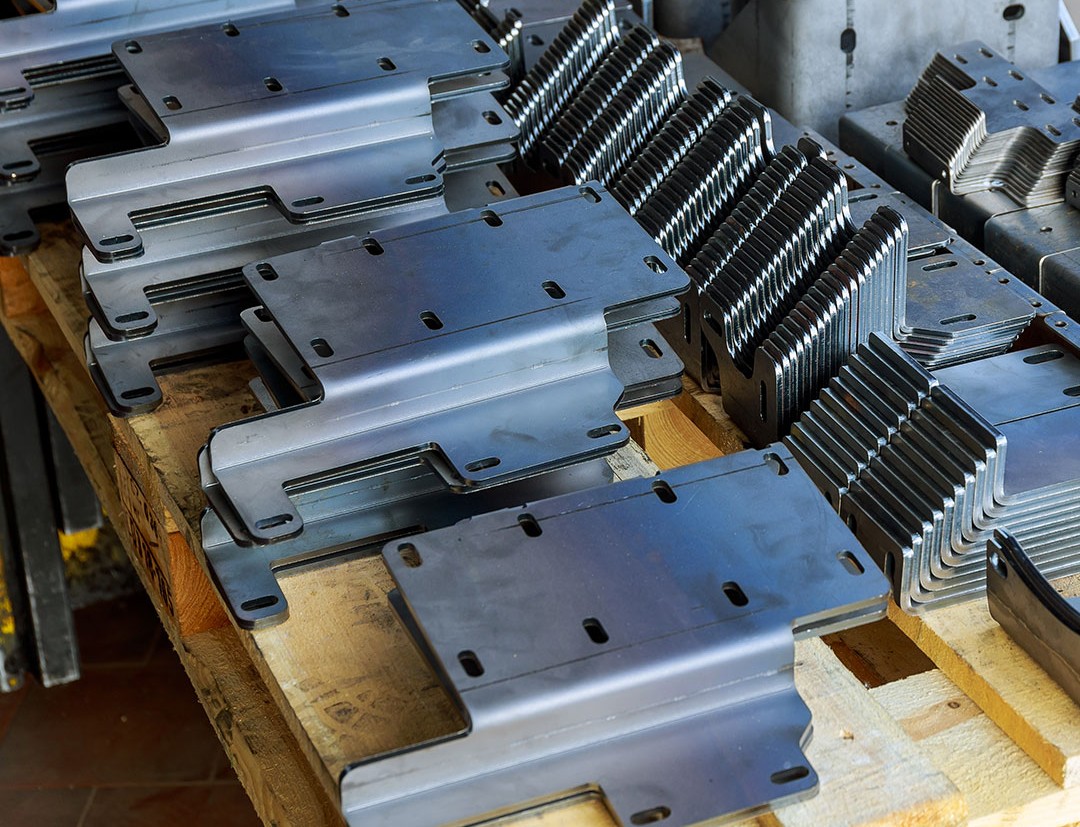}
     \end{subfigure}
    \caption{Unstructured supply in a disordered bin~\cite{WiredWorkers:2025} (top); semi-structured stacks of products~\cite{JanSteel:2025} (bottom).}\label{fig:less_structured_part_supplies}
\end{figure}

\subsection{Dealing with (upstream) errors}

On the other hand, in case parts' poses are fixed or accurately known beforehand, object recognition and localisation could allow for process verification, that is, to check whether the cobot is provided with the right part (object recognition) in the correct position and/or orientation (pose estimation). A discrepancy may stem from a malfunctioning machine or process step upstream in the production line. In such case, the cobotic assistant could compensate for it by incorporating the estimated pose of the workpiece in its motion planning. When a wrong product is recognised, the cobot could alert the process supervisor to prevent greater harm. Recording these data enables workshop owners in making proactive corrections to the (semi-automated) manufacturing process, leveraging the abnormalities identified by the cobotic production assistant.

\subsection{Contactless referencing}\label{sec:contactless_referencing}

\begin{figure}
	\centering
	\includegraphics[width=0.9\linewidth]{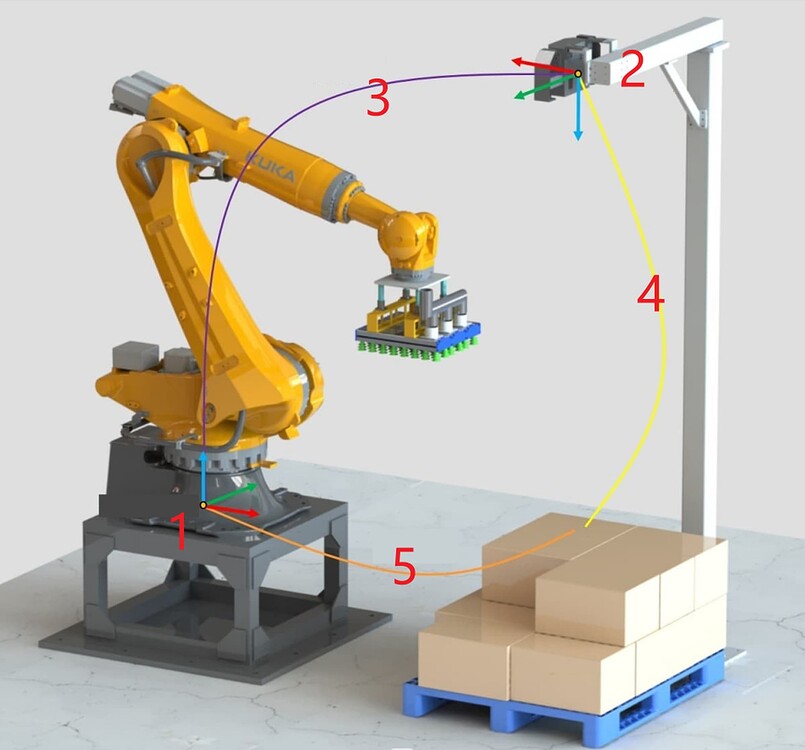}
	\caption{Visual representation of coordinate transformations between the world, the robot and a static camera~\cite{MechMind:2023}.}\label{fig:coordinate_transformations_robot_setup}
\end{figure}

Finally, pose estimation could be adopted to determine the cobot's pose relative to the elements of its environment it must interact with, e.g., machines, mobile part racks, without physical connections (see Figure~\ref{fig:coordinate_transformations_robot_setup}). For example, by firmly attaching a dedicated reference object to a machine or by even using a distinctive part of the machine itself, the CAD model of this object or machine part could be fed to a 3D matching algorithm calculating the pose of the cobotic workstation with respect to the production machine.

Contactless relative positioning eliminates the necessity to layout the workshop with intrusive anchoring points, often permanently secured to a wall or floor, for a mobile workstation to dock with. Moreover, in specific circumstances, a contact-based relative positioning between setups alone does not suffice. Insights from industry show that severe vibrations over extended periods of time can wear the damping elements of, for instance, a robotic workstation or machine. This deterioration has a degressive effect on the position and orientation of the entire arrangement, ultimately requiring the relative positioning procedure to be renewed. Vision-based referencing, on the other hand, could be regularly repeated in an automated manner, as part of the overall workflow, without the need for human intervention.

%------------------------------------------------

\section{Industrial use case}

Within the context of a past feasibility study, ACRO implemented and compared several of HALCON's 3D matching algorithms for the robotic handling of sheet-metal products. The objective consisted of extracting the 6D poses of laser-cut workpieces from point cloud data, enabling a robotic manipulator to accurately grasp them for further processing. While CAD models of the products were available, no prior information was provided regarding their position, orientation, or type. The workpieces were presented to the robot in either unstructured or semi-organised arrangements. Additionally, the client specified the industrial depth camera to be used for this project.

\subsection{Challenges}\label{sec:challenges}

\begin{enumerate}[label=\Roman*]
\item \emph{Various input streams}: semi-finished products can be presented to the robot on roller conveyors, stacked on wooden or steel pallets, and positioned on a punching machine's press bed. Hence, the developed computer vision software must be robust against changing and cluttered backgrounds.

\item \emph{Mixed inputs}: the workpieces supplied can be a random assortment of 30 different variants. Which specific product type follows next is unknown. Therefore, the developed computer vision software must be capable of identifying products in addition to determining their poses.

\item \emph{Sheet-metal parts}: the products to be handled are relatively thin with respect to their other dimensions. Since there are little to no features available along the Z-axis, these workpieces approximate 2D objects. Nonetheless, their poses must be determined in 6D. Additionally, the relatively low stiffness and thickness of these sheet-metal parts allow for deflections\footnote{The interested reader is referred to the \emph{deformable} surface-based matching algorithm offered by HALCON. It allows localising objects prone to bending by generating deformable states from the provided CAD model.}, as they are only partially supported by the conveyor's widely spaced rollers.

\item \emph{Industrial environment}: the robot operates within an industrial environment where conditions are far from ideal for computer vision systems. Due to the presence of a considerable amount of incident natural light, ambient lighting conditions fluctuate throughout the day. Moreover, infrared light is known to interfere with certain depth cameras, especially when combined with the reflective, curved surfaces of the conveyor's rollers. These rollers may affect the camera or sensor's own light source (if any) in a similar, adverse way.

\item \emph{Different-sized products}: the dimensions of a semi-finished product can range from about \SI{30}{\centi\meter} up to \SI{2}{\meter}. In case of the latter, the workpiece may be partially outside the camera's field of view. The developed computer vision software must, therefore, be capable of operating with only the product segment(s) visible in the camera data or, alternatively, support the merging of multiple depth images or point clouds into a unified view\footnote{This latter option is also known as \emph{point cloud stitching} or \emph{image mosaicking}~\cite[Ch. 9]{MVTec:2025a} depending on the type of data that must be merged.}.

\item \emph{Cycle time}: since the software's performance is not only measured by its accuracy but also by its processing speed, detecting a single workpiece must occur within a timeframe of just a few seconds. Of course, the faster the better applies here. As point clouds typically contain more data than a colour image of the same scene, complying with this requirement is not as straightforward as it may seem.
\end{enumerate}

\subsection{Robot scanning configuration}

\begin{figure}[htb]
	\centering
	\includegraphics[width=0.7\linewidth]{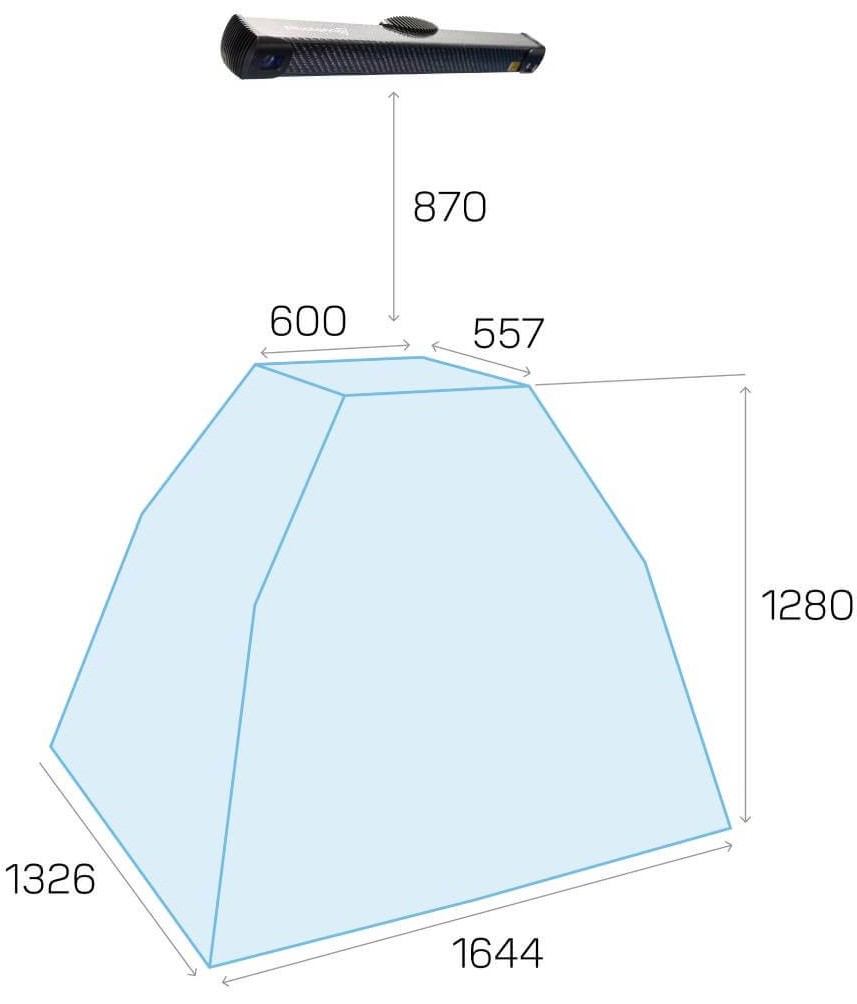}
	\caption{Scanning volume of the Photoneo PhoXi 3D scanner L~\cite{Photoneo:2025}.}\label{fig:phoxi_3d_scanning_volume}
\end{figure}

For this project, the camera is attached to the robot's tool flange, next to its gripper: a setup referred to as \emph{eye-in-hand}. To identify a camera pose that comprehensively and accurately captures the product, regardless of its potential positions and orientations, the robot had to be manually jogged through as much of its workspace as possible. Alternatively, potentially optimal scanning poses can first be explored in simulation, allowing broad coverage and rapid iteration. These candidates must then be validated on the physical setup to confirm their viability in the real world. The result should consist of at least one optimal scanning configuration, determined by considering aspects such as:
\vspace{1ex}

\textbf{Is the workpiece within the scanning range of the camera used?} Every depth camera comes with a minimum and maximum scanning distance (see Figure~\ref{fig:phoxi_3d_scanning_volume}). Only elements in the scene that lie within this distance range, with respect to the camera, are captured. Preferably, when specified by the camera manufacturer, robot configurations that position the camera at its optimal scanning distance (``sweet spot'') should be explored first.

\textbf{Is the product entirely visible in the depth data?} For many of HALCON's 3D matching algorithms to perform accurately, the depth data should contain the complete workpiece. This is especially true for those techniques that rely on a CAD model and try to match it entirely in the provided data. When capturing the whole workpiece in a single shot is not practical---for instance, if the product is too large or the camera is positioned too close---an alternative approach could involve merging multiple point clouds or dividing the CAD model into smaller segments. Additionally, one should ensure that the product occupies the maximum possible area within the camera's scanning range, as this increases the number of pixels representing the surface of the object within the depth image---commonly referred to as the \textit{pixel density} over the object. Higher pixel density leads to richer detail, more visible features, reduced influence from irrelevant background, and lower measurement noise, all of which contribute to more accurate and reliable 3D object localisation and recognition. Moreover, it is important to check that the entire product remains visible, even when its orientation or position changes. A recommended approach is to place the product in its most extreme poses and verify whether the camera can still capture it effectively.
\vspace{1ex}

\begin{figure}
	\centering
	\begin{subfigure}[b]{0.465\textwidth}
		\includegraphics[width=\textwidth]{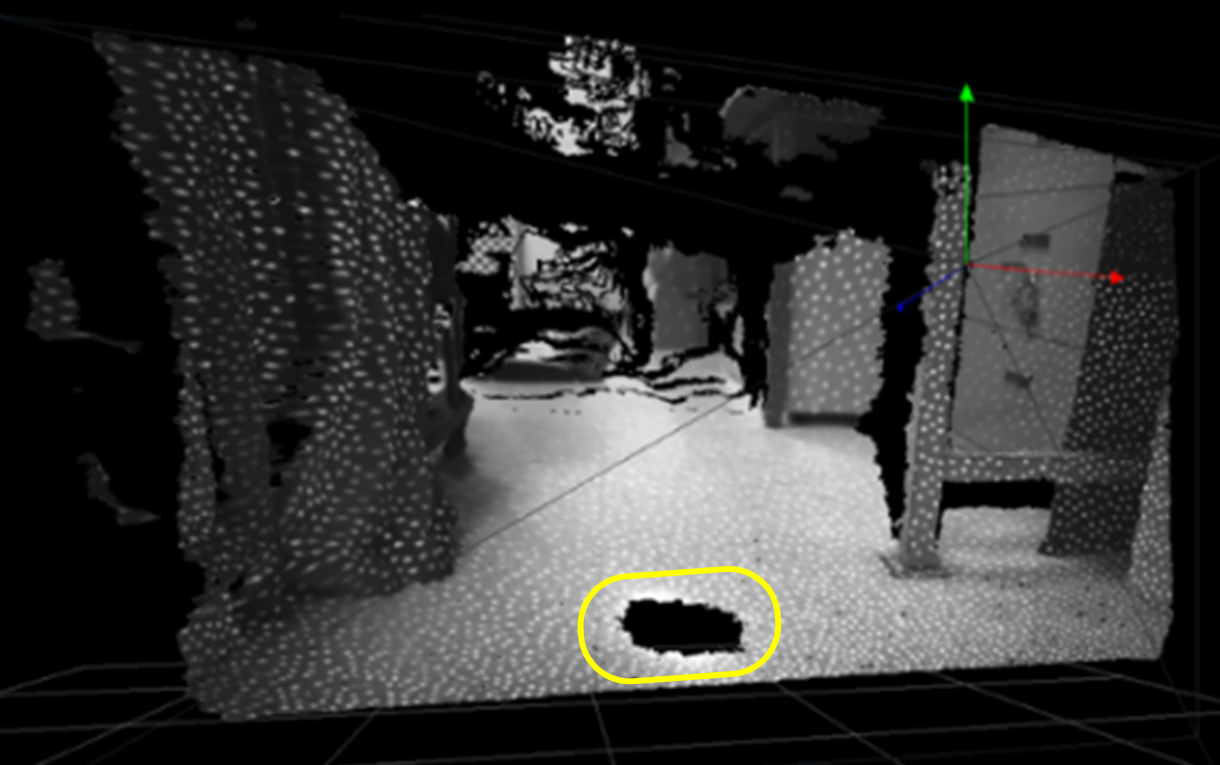}
	\end{subfigure}
	\par\smallskip
	\begin{subfigure}[b]{0.465\textwidth}
		\includegraphics[width=\textwidth]{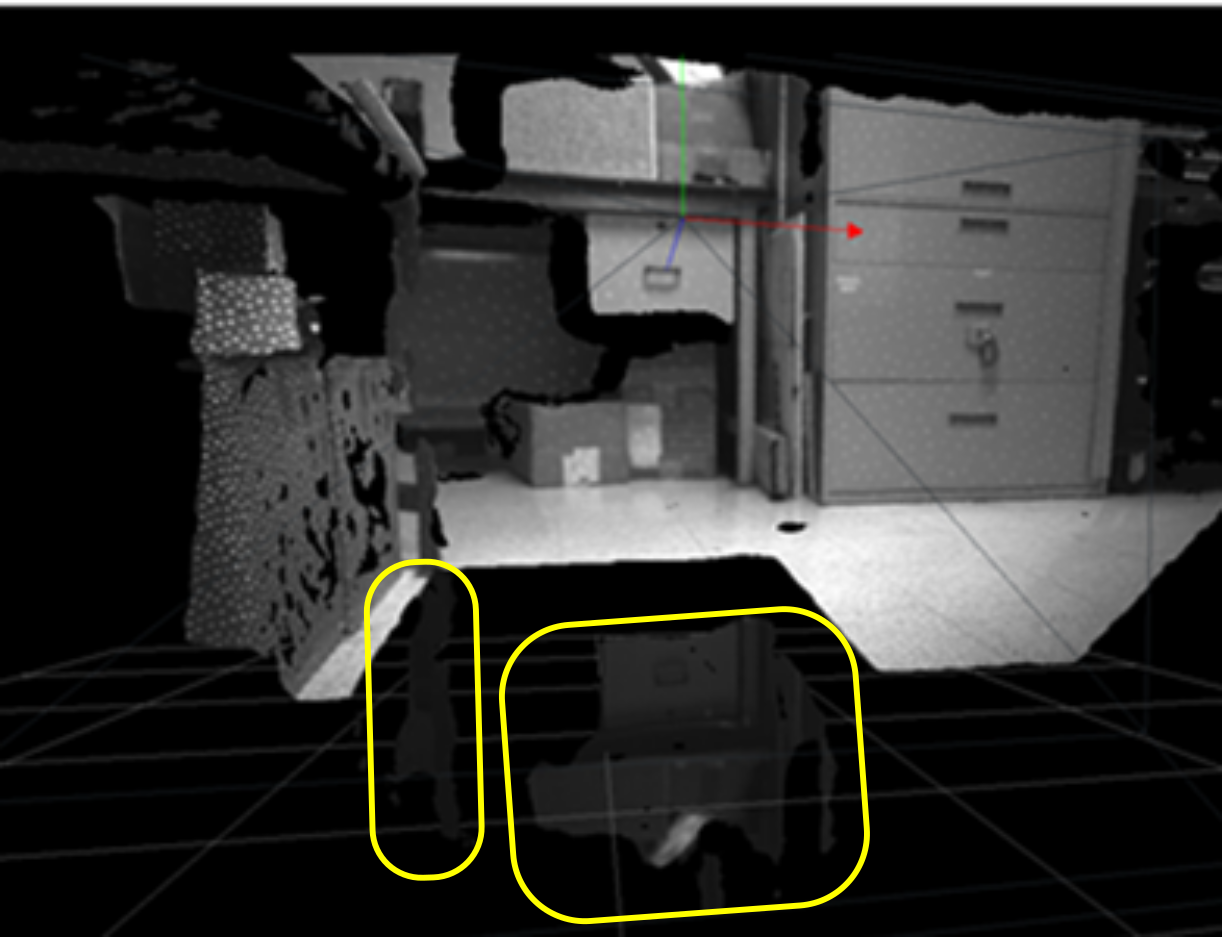}
	\end{subfigure}
	\caption{Point cloud defects such as holes (top) and false artifacts (bottom) caused by local saturation and specular reflections~\cite{Sweetser:2023}.}\label{fig:point_cloud_defects}
\end{figure}

\textbf{Are there optical disturbances present, linked to the camera's current pose, that can be avoided?} When observing clusters of pixels in the monochrome depth image being affected by bright incident light or glare, adjusting the camera angle by moving the robot arm may sometimes help mitigate the issue~\cite{Photoneo:2019}. The local saturation of specific pixels of the depth camera's sensor(s)---whether induced by specular reflections---can namely result in considerable data loss, such as holes in a point cloud (see Figure~\ref{fig:point_cloud_defects}, top). Specular reflections can also produce the opposite effect, generating false depth artifacts i.e., objects in the point cloud that are not physically present (see Figure~\ref{fig:point_cloud_defects}, bottom). However, sometimes, adjusting the camera's pose may not be enough, requiring modifications to its settings or the integration of external optical components, such as filters and polarisers~\cite{Sweetser:2023}. One must finally note that the sensitivity of a depth camera to certain optical disturbances also largely depends on its operating principle.

\subsection{Point cloud refinement and surface-based matching}

\begin{figure}[htb]
	\centering
	\includegraphics[width=\linewidth]{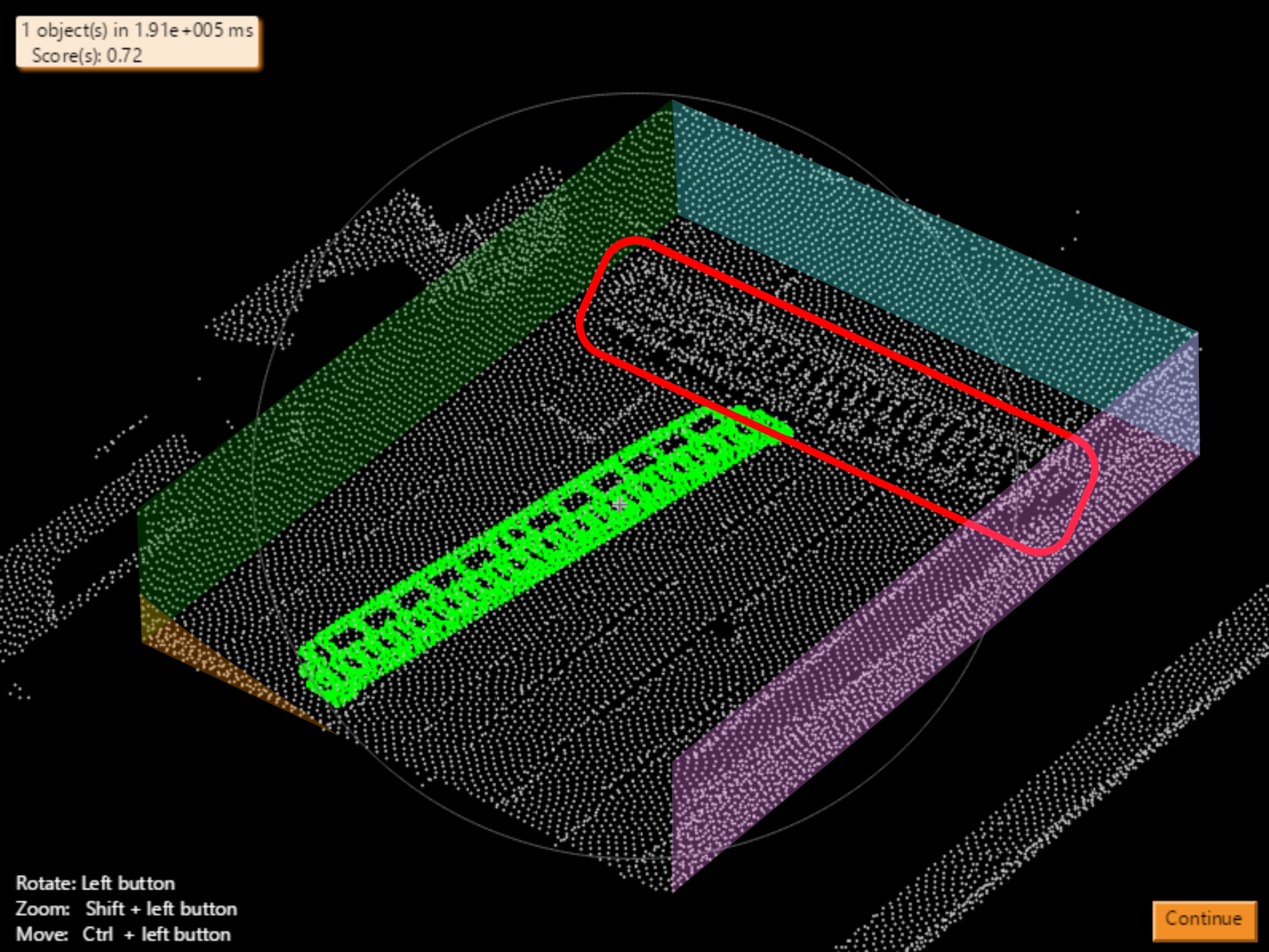}
	\caption{Raw point cloud of a framed pallet with a single workpiece. Coloured surfaces mark the pallet's side planks; points within the red rectangle indicate the workpiece. The bright green CAD model shows the result of 3D surface-based matching on the raw point cloud, revealing a visible pose mismatch. Matching completed in \num{191} seconds with a score of \num{0.72}.}\label{fig:surface_based_matching_raw_point_cloud}
\end{figure}

Once the optimal robot scanning configuration is established and depth camera data is acquired, the next step is to determine the 6D pose of the product. Figure~\ref{fig:surface_based_matching_raw_point_cloud} pictures the outcome of estimating the product's pose by applying the 3D surface-based matching algorithm to the raw point cloud. It is evident that the estimated pose of the workpiece, represented by the green point-sampled CAD model, deviates significantly from its actual position and orientation within the red rectangle. This discrepancy may be due to the matching algorithm becoming ``distracted'' by the narrow boards of the framed pallet in Figure~\ref{fig:monochrome_image}, which share a slight resemblance to the elongated, thin product\footnote{The matching score for this test was \num{0.72} out of \num{1.0}, indicating how closely the algorithm erroneously perceived the match to be perfect. The computation time was \SI{3}{\minute} and \SI{11}{\second}, which is unacceptably long.}.

As previously mentioned, refining a point cloud by removing data points associated with the scene's background leads to quicker and more precise pose estimations. Various attributes of a data point (or a set of data points) could assist in differentiating between the background and the workpiece.

\subsubsection{Distance with respect to the camera}\label{sec:z_channel_thresholding}

\begin{figure}[htb]
	\centering
	\includegraphics[width=\linewidth]{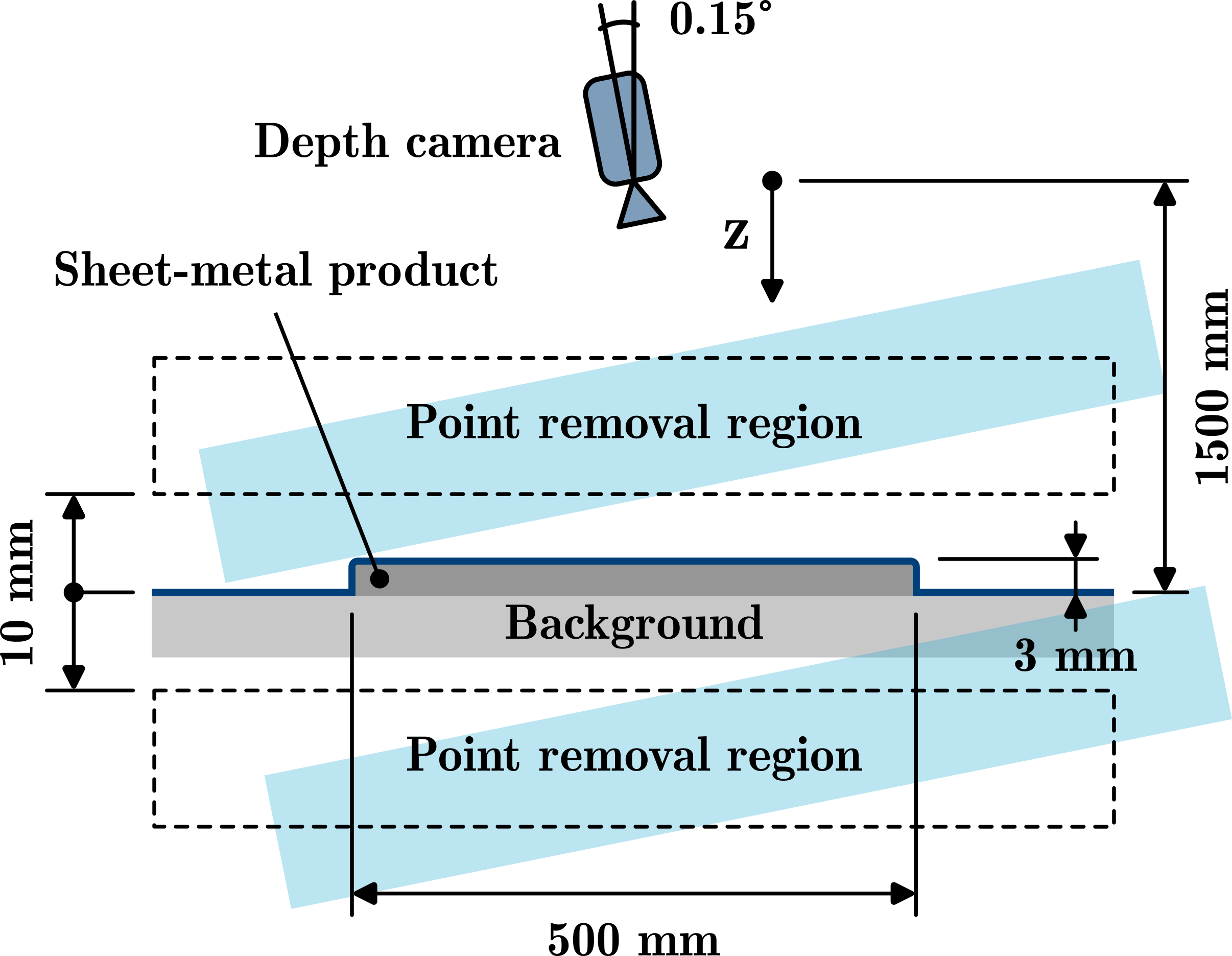}
	\caption{A depth camera positioned \SI{1500}{\milli\meter} from the scene captures a \SI{3}{\milli\meter} thick, \SI{500}{\milli\meter} long product. A minor camera tilt inaccuracy of just \SI{0.15}{\degree} can lead to the exclusion of product points when Z-channel thresholding is applied with a \SI{10}{\milli\meter} threshold range. This highlights the need for precise extrinsic calibration.}\label{fig:camera_tilting_accuracy_z_thresholding}
\end{figure}

By definition, the background is positioned farther from the camera than the foreground. Z-channel thresholding involves filtering out all points whose distance from the camera, along the Z-axis of its coordinate frame, falls outside a specified threshold range. For optimal performance, the scene's normal vector (e.g., the framed pallet) should be aligned with the camera's Z-axis. The best practice is to determine the coordinate transformation between the camera and the scene through calibration, allowing Z-thresholding to be applied in the scene's coordinate system. As illustrated in Figure~\ref{fig:camera_tilting_accuracy_z_thresholding}, even a minor camera tilt inaccuracy of \SI{0.15}{\degree} can cause Z-thresholding to remove point cloud data belonging to the product, highlighting the need for precise extrinsic calibration.

\begin{figure}[htb]
	\centering
	\includegraphics[width=\linewidth]{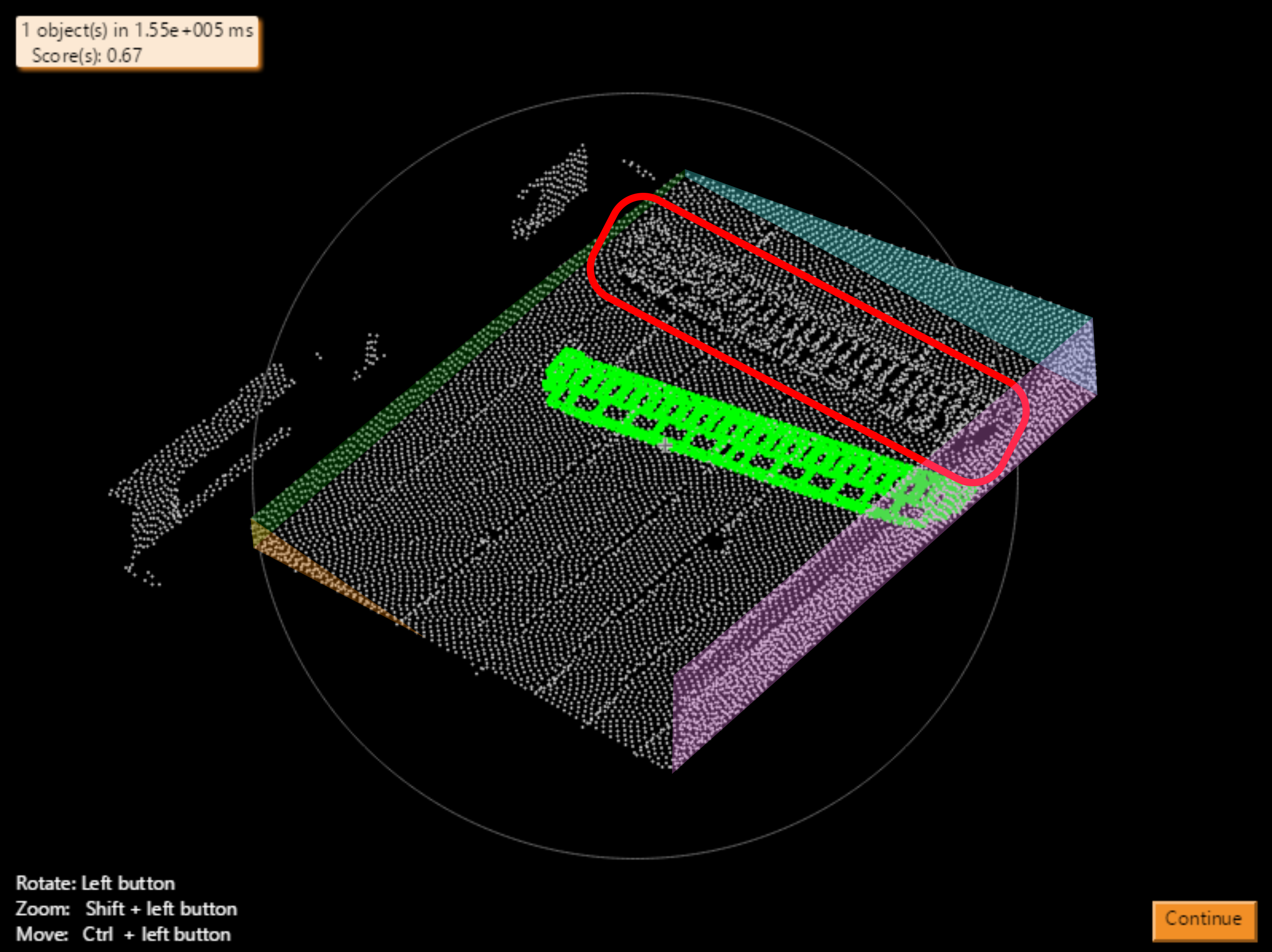}
	\caption{Reduced point cloud of a framed pallet with a single workpiece. Coloured surfaces mark the pallet's side planks; points within the red rectangle indicate the workpiece. The bright green CAD model shows the result of 3D surface-based matching after applying Z-channel thresholding to the raw point cloud, revealing a visible pose mismatch. Matching completed in \num{155} seconds with a score of \num{0.67}.}\label{fig:surface_based_matching_z_thresholding}
\end{figure}

Because the extrinsic parameters between the camera and the scene were initially unknown, applying thresholds according to the Z-axis of the camera led to the near-total removal of the side plank on one side of the pallet, while the board on the opposite side remained largely intact (see Figure~\ref{fig:surface_based_matching_z_thresholding}). Consequently, the workpiece was incorrectly matched once again.

\subsubsection{Grey value in the monochrome image}

\begin{figure}[htb]
	\centering
	\includegraphics[width=0.95\linewidth]{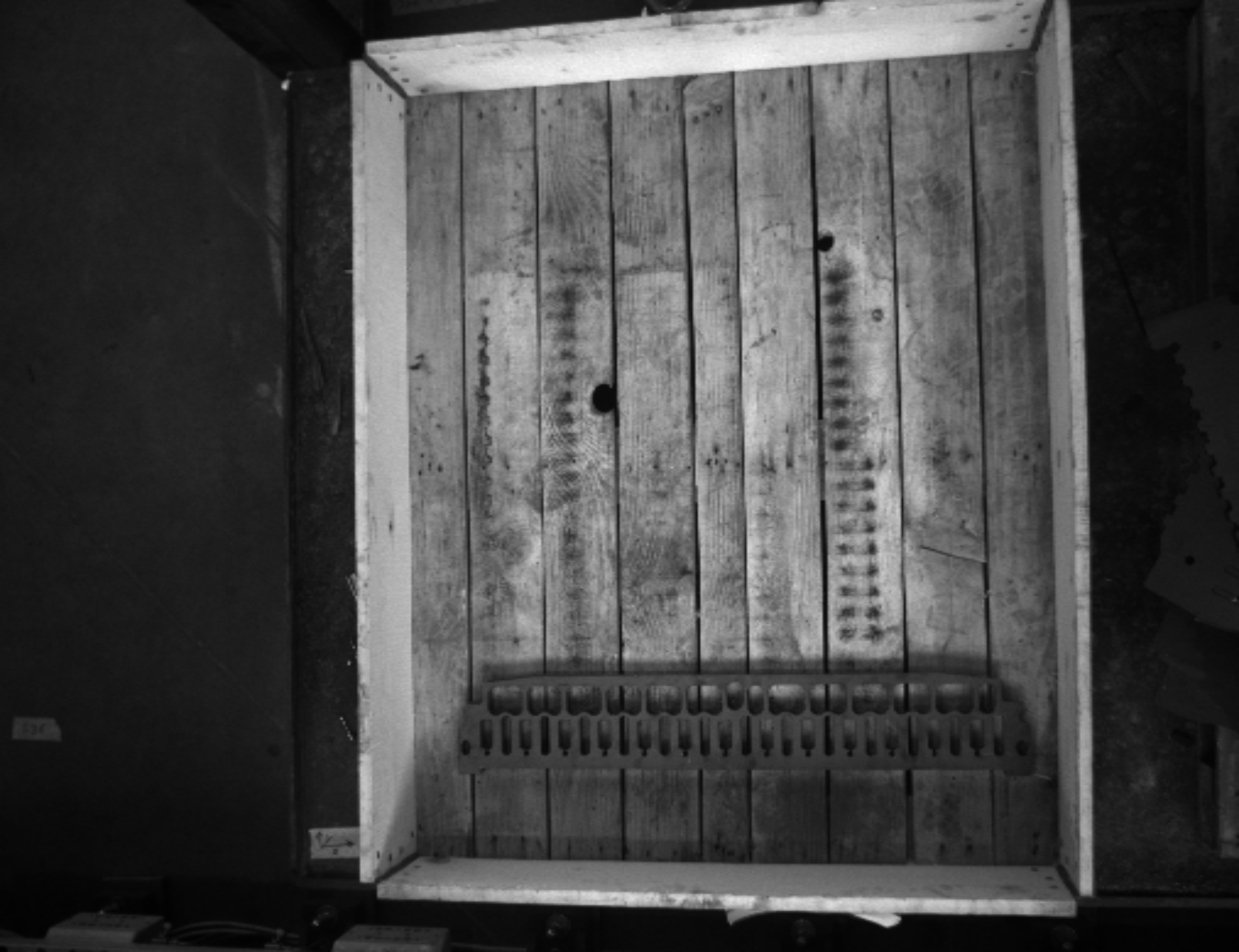}
	\caption{Monochrome image of a framed pallet with a single workpiece in an industrial setting.}\label{fig:monochrome_image}
\end{figure}

\begin{figure}[htb]
	\centering
	\includegraphics[width=\linewidth]{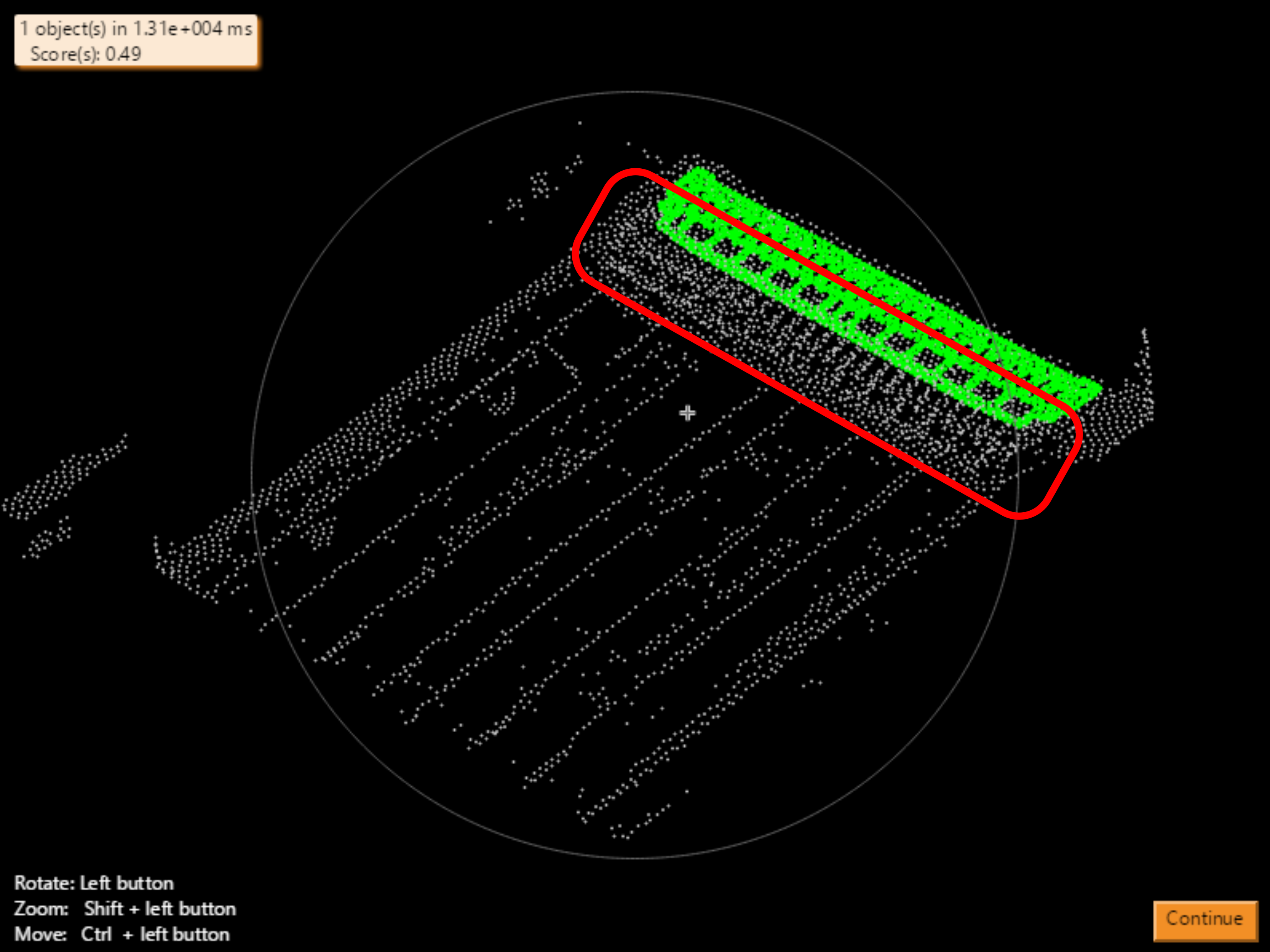}
	\caption{Reduced point cloud of a framed pallet with a single workpiece. Points within the red rectangle indicate the workpiece. The bright green CAD model shows the result of 3D surface-based matching after applying greyscale thresholding (cfr. Figure~\ref{fig:monochrome_image}) to the coloured raw point cloud, revealing a visible pose mismatch. Matching completed in \num{13.1} seconds with a score of \num{0.49}.}\label{fig:surface_based_matching_greyscale_thresholding}
\end{figure}

At times, the product and its background can be visually distinct, characterised by their different colours. Figure~\ref{fig:monochrome_image}, for instance, shows a greyscale image of a scene. The significant contrast between the light wooden planks and the dark steel product enabled the removal of a large portion of the background using a threshold to the pixels' grey values (see Figure~\ref{fig:surface_based_matching_greyscale_thresholding}). However, shadows and grime created a dark area at the bottom of the image with grey values closely resembling those of the product. This led to a dense cluster of points in the point cloud, which still prevented a correct matching. Given that a high contrast between the product and background cannot always be ensured due to unpredictable lighting conditions, this filtering technique does not always yield satisfactory results. If the 3D camera is also equipped with a colour camera, a similar filter could be applied to the three RGB channels. Unfortunately, experience shows that colour-based filtering is even more susceptible to variations in ambient lighting.

\subsubsection{Orientation of the normal vector}

\begin{figure}[htb]
	\centering
	\includegraphics[width=\linewidth]{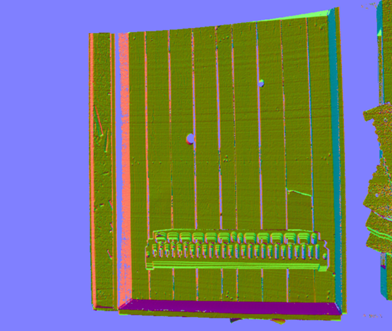}
	\caption{Normal map of a framed pallet with a stack of workpieces. Its distinct colours represent clusters of normal vector orientations.}\label{fig:normal_map}
\end{figure}

\begin{figure}[htb]
	\centering
	\includegraphics[width=\linewidth]{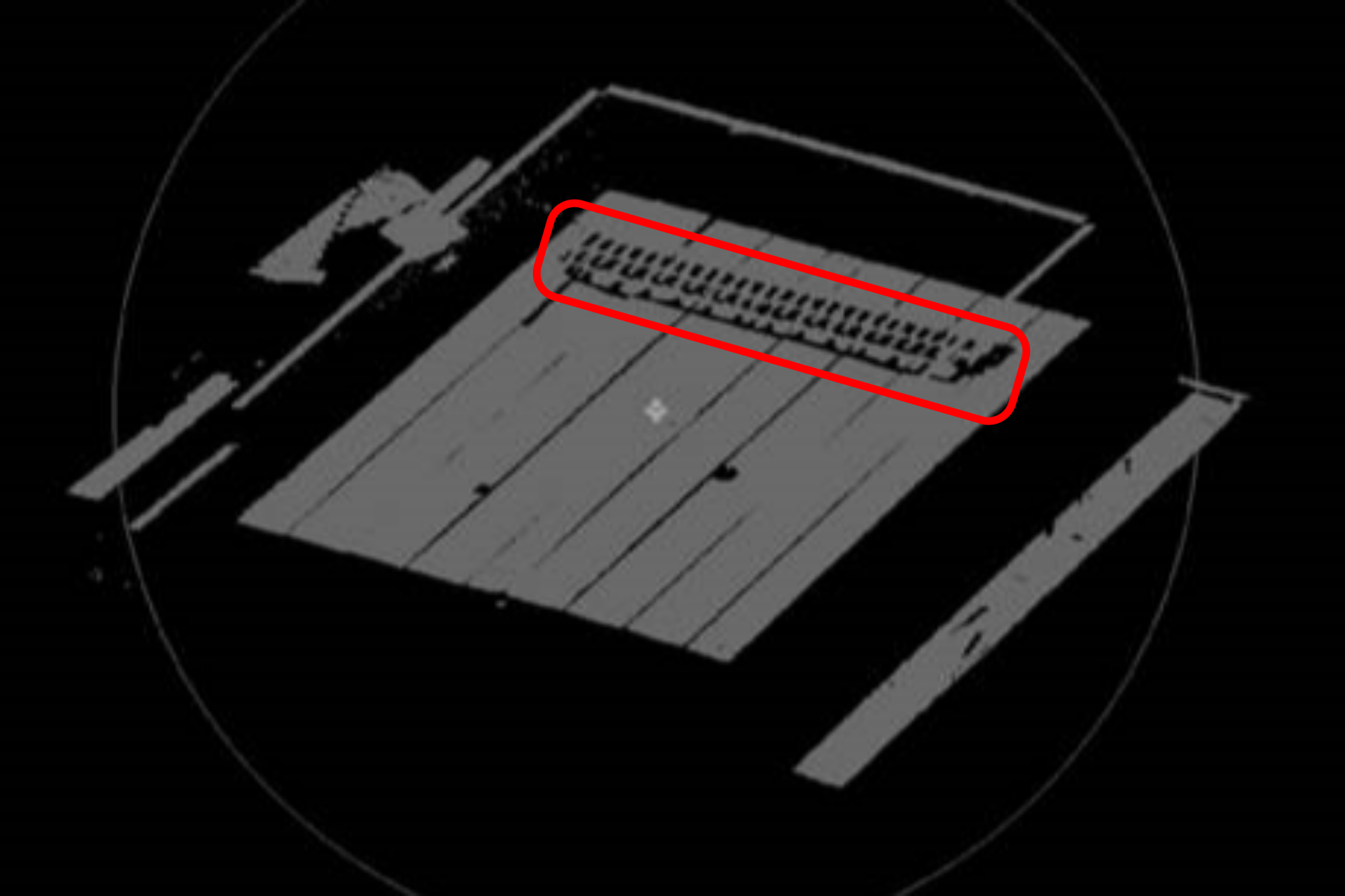}
	\caption{Reduced point cloud of a framed pallet with a single workpiece, obtained by removing points belonging to (nearly) vertical surfaces such as the pallet's side planks. Points within the red rectangle indicate the workpiece.}\label{fig:normal_vector_thresholding}
\end{figure}

If the scene includes elements, such as machine panels or walls, with orientations different from those of the workpieces, the corresponding data points can be isolated by analysing their normal vectors' directions. Applied to our current use case, the pallet's side planks could be filtered out from the point cloud as their surfaces' normal vectors are perpendicular to those of the bottom planks and the product. Figure~\ref{fig:normal_map} depicts the normal map outputted by the depth camera, where its discrete colour scale relates to different ranges (clusters) of normal directions. The dark green points have normal directions corresponding to that of the product and are the ones targeted for isolation. By applying this type of filter, the point cloud of Figure~\ref{fig:normal_vector_thresholding} is obtained with only points remaining that belong to horizontal planes.

\subsubsection{Presence of flat surfaces}

\begin{figure}[htb]
	\centering
	\includegraphics[width=\linewidth]{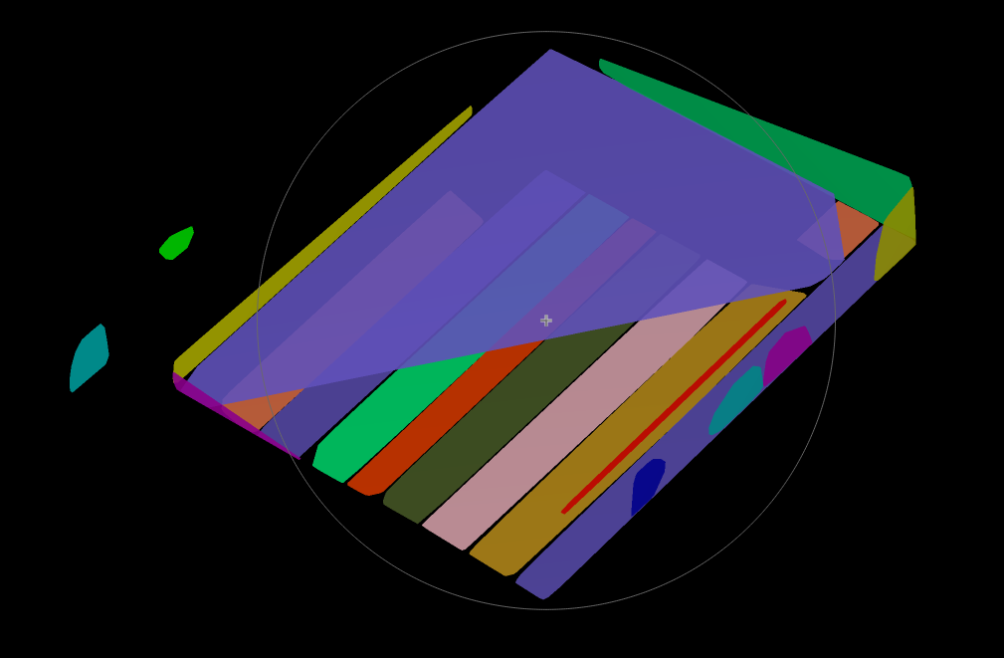}
	\caption{Planar segmentation of the reduced point cloud shown in Figure~\ref{fig:surface_based_matching_z_thresholding}, performed using 3D primitives fitting.}\label{fig:plane_segmentation}
\end{figure}

While the previous approach could be effective in eliminating depth points from various---including irregularly curved---surfaces, man-made environments are typically dominated by flat surfaces such as walls, tabletops, and machine panels. Hence, clusters of 3D points associated with these surfaces can be segmented by fitting 2D planes through them. In the presented use case, the bottom planks and side slats of the framed pallet align with this criterion. Beyond planes, the 3D primitive fitting algorithm in HALCON can also fit other geometric shapes, such as cylinders and spheres, to a set of 3D points. The result of applying this algorithm is pictured in Figure~\ref{fig:plane_segmentation}. Subsequently, the distances between each primitive plane---excluding those intersecting the product of interest---and the 3D points are determined. All points within a defined range from these planes are removed, as they are considered part of an irrelevant flat artifact in the environment.

\begin{figure}[htb]
	\centering
	\includegraphics[width=\linewidth]{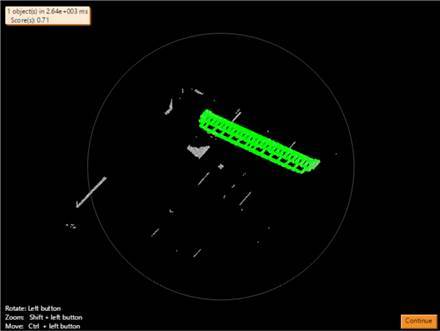}
	\caption{Reduced point cloud of a framed pallet with a single workpiece. The bright green CAD model indicates a successful 3D surface-based matching, performed after background subtraction using 3D primitive fitting, with no visible pose mismatch. Matching completed in \num{2.64} seconds with a score of \num{0.71}.}\label{fig:surface_based_matching_primitives_fitting}
\end{figure}

Figure~\ref{fig:surface_based_matching_primitives_fitting} illustrates how this method effectively isolates 3D points associated with the product, leading to a precise and rapid matching of the CAD model. Indeed, reducing the point cloud accelerated the 3D surface-based matching process significantly. On the flip side of the coin, however, is the relatively high computation time (\(\pm\) \num{15} to \SI{20}{\second}) required for plane segmentation and background subtraction. If multiple products must be retrieved from the same pallet, the fitted planes can potentially be reused for other products as well, allowing this computation time to be distributed across all items on the pallet.

When applying 3D primitives fitting to the reduced point cloud shown in Figure~\ref{fig:normal_vector_thresholding} instead of the raw data, a time savings of approximately \SI{10}{\second} is achieved due to the absence of the side planks. Down sampling the point cloud could also help accelerating the pre-processing stage.

\subsubsection{Presence of static structures}

\begin{figure}[htb]
	\centering
	\includegraphics[width=\linewidth]{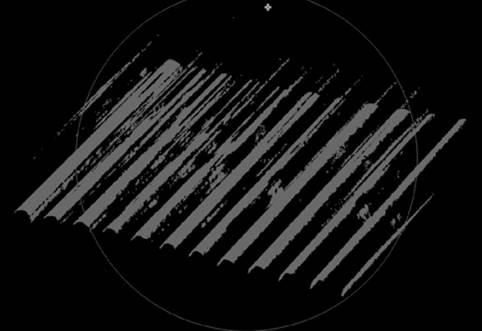}
	\caption{Reference point cloud of an empty roller conveyor.}\label{fig:reference_point_cloud}
\end{figure}

In many industrial settings, the position of elements like walls, floors, conveyor belts, and machines remains unchanged for extended periods of time. If the camera pose---and therewith the robot's scanning configuration---also stays fixed, the depth data associated with these static structures can be eliminated from the raw point cloud by ``subtracting'' it with a reference point cloud\footnote{This operation should be performed with a tolerance margin, as slight deviations in the robot's scanning pose and minor fluctuations in background measurements may occur.}  that excludes all desired objects, such as the sheet-metal workpieces. Figure~\ref{fig:reference_point_cloud} pictures the reference point cloud of an empty roller conveyor, used in the presented use case.

\begin{figure}[htb]
	\centering
	\includegraphics[width=\linewidth]{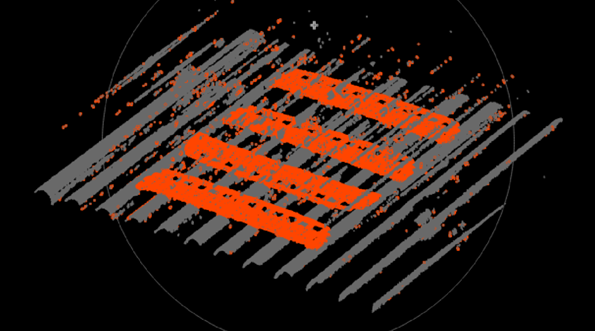}
	\caption{Raw point cloud (grey) superimposed with the set of points (orange) remaining after the comparison operation with the reference point cloud in Figure~\ref{fig:reference_point_cloud}.}\label{fig:subtracted_point_cloud}
\end{figure}

The orange point cloud in Figure~\ref{fig:subtracted_point_cloud} shows the remaining data after performing this comparison operation. First, the distances between the points in both the raw and reference point clouds are computed. Next, only the 3D points\footnote{Depending on the camera's operating principle, and consequently the availability of depth images/maps, the comparison operation may be performed directly on depth images, eliminating the need for conversion into point clouds.} that have no nearby neighbour in the reference point cloud are preserved. The threshold value that determines the maximum inner point distance at which an adjacent point is considered close, can be set in the corresponding HALCON function.

\begin{figure}[htb]
	\centering
	\includegraphics[width=\linewidth]{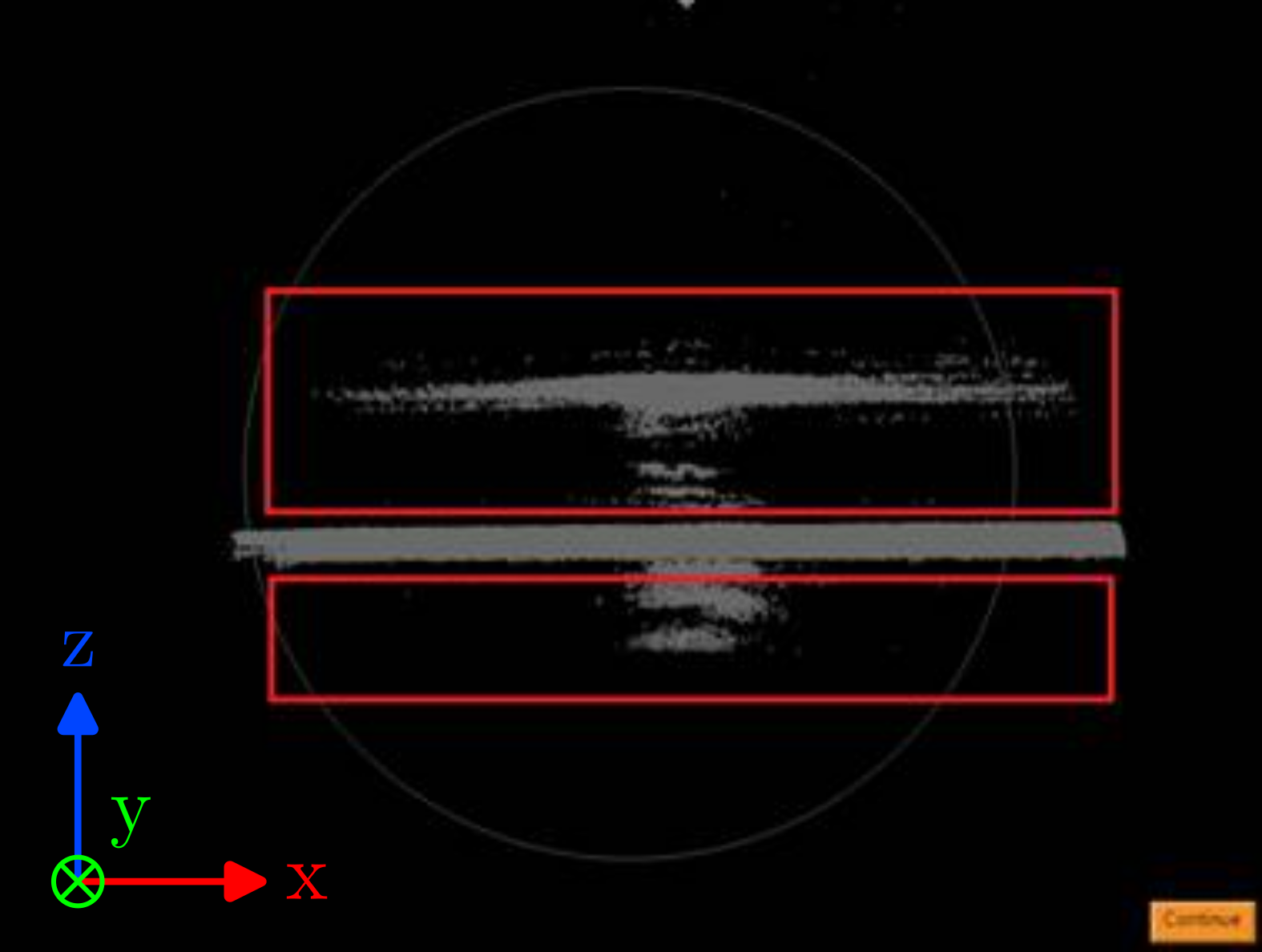}
	\caption{Candidate boundaries for eliminating the ghost points above and below the roller conveyor's raw point cloud using Z-channel thresholding.}\label{fig:z_thresholding_ghost_points}
\end{figure}

The presence of floating artifacts above the roller conveyor in both figures indicates that the depth camera encountered difficulties in accurately capturing the reference scene. This challenge arose because the adopted high-precision camera relied on structured-light projection to determine depth. By analysing distortions in these projected geometric patterns, it calculates the 3D shape of objects. However, the reflective nature of the rollers, combined with their curved surfaces, caused the projected (red laser) light to either bounce back to the camera at unintended angles or undergo multiple reflections between adjacent rollers. As a result, the depth camera misinterpreted these reflections as misplaced or duplicate surfaces, producing ghost points within the scene. Because these reflections can vary slightly between captures, ghost points may not always appear in the same locations, allowing some to evade the comparison operation. If the coordinate transformation between the camera and the roller conveyor is known, setting a Z-threshold slightly above the workpieces could help eliminating them (see Figure~\ref{fig:z_thresholding_ghost_points}). As some of these point clusters are small in size and isolated, a statistical outlier filter~\cite{PCL:2025} could additionally help in pre-processing the raw point cloud for the subsequent 3D matching stage.

\subsubsection{Sharp depth gradients}

There have been multiple instances where objects were incorrectly localised in the scene's background (see Figures~\ref{fig:surface_based_matching_raw_point_cloud},~\ref{fig:surface_based_matching_z_thresholding},~\ref{fig:surface_based_matching_greyscale_thresholding}) due to its surface roughly matching the shape of the workpieces. Thus, it appears that the surface-based matching algorithm disregards the product's prominent contours, mistakenly treating it as part of a smooth, uniform background. HALCON's edge-supported surface-based matching specifically targets to address this need for integrating 3D edge data into the pose estimation process. It identifies sharp variations or gradients within the point cloud and attempts to align them with the edges of the CAD model.

\begin{figure}[htb]
	\centering
	\begin{subfigure}[b]{0.45\textwidth}
		\includegraphics[width=\textwidth]{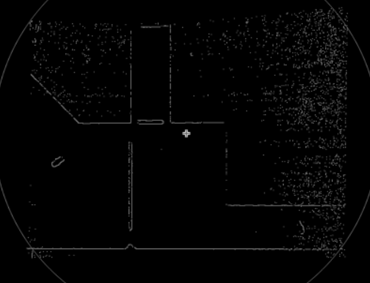}
	\end{subfigure}
	\par\smallskip
	\begin{subfigure}[b]{0.45\textwidth}
		\includegraphics[width=\textwidth]{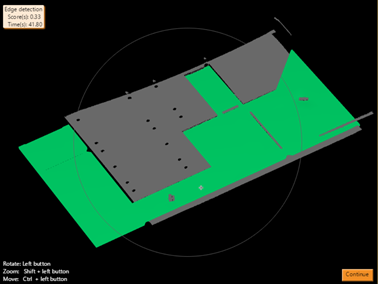}
	\end{subfigure}
	\caption{Point cloud containing only edge points extracted from the raw data by the 3D edge detector (top). Raw point cloud of a solid deck steel pallet with a partially visible large product. The green CAD model shows the moderately successful result of 3D edge-supported surface-based matching (bottom). Matching completed in \num{41.8} seconds with a score of \num{0.33}.}\label{fig:edge_supported_surface_based_matching}
\end{figure}

Figure~\ref{fig:edge_supported_surface_based_matching} (top) visualises the output of the 3D edge detector: a point cloud encoding the 3D coordinates and viewing directions of all edges extracted from the raw depth data. However, in addition to the product's contours, noise appears along the top and right borders of this image. This may originate from carves, crevices, or minor height variations in the background, being misidentified as edges. To mitigate this, the minimum amplitude at which a discontinuity in the depth data is classified as an edge can be adjusted. In the presented use case, this threshold needed to align with the thickness of the relatively thin sheet metal, causing unwanted background discontinuities to become more prominent. Hence, this 3D matching technique would yield even better results for taller products that extend significantly further into the foreground, closer to the camera.

Despite the presence of noise in the edge data and the sizeable product being only partially captured by the depth camera, the edge-supported surface-based matching of the CAD model (green) within the point cloud (grey) was relatively accurate as shown in Figure~\ref{fig:edge_supported_surface_based_matching} (bottom).   

\subsection{Shape-based matching}\label{sec:shape_based_matching}

\begin{figure}[htb]
	\centering
	\begin{subfigure}[b]{0.48\textwidth}
		\includegraphics[width=\textwidth]{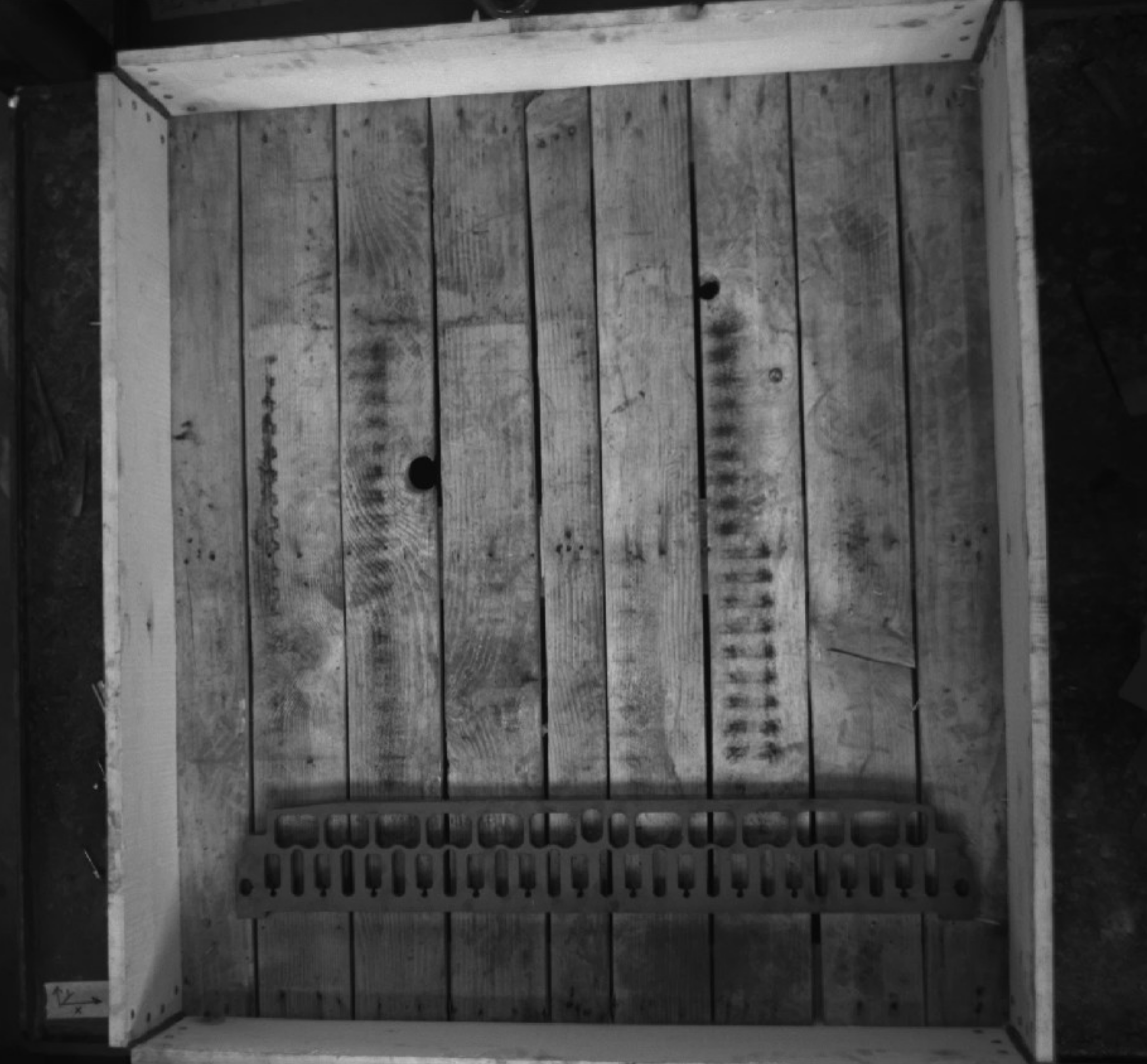}
	\end{subfigure}
	\par\smallskip
	\begin{subfigure}[b]{0.48\textwidth}
		\includegraphics[width=\textwidth]{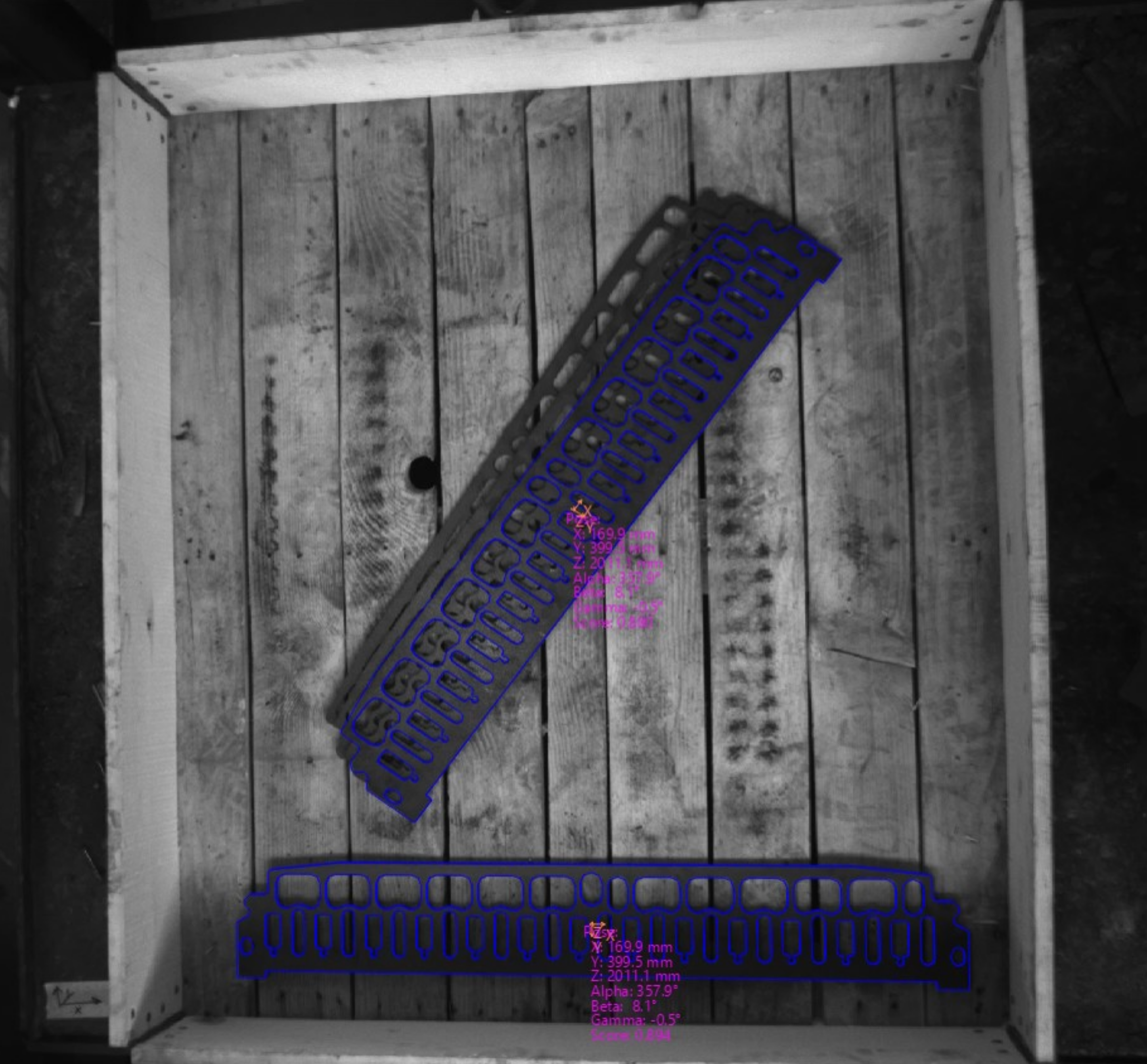}
	\end{subfigure}
	\caption{Greyscale images of framed pallets with stacks of workpieces where 3D shape-based matching failed (top) and succeeded (bottom).}\label{fig:shape_based_matching_greyscale}
\end{figure}

As mentioned in Section~\ref{sec:terminology}, 3D shape-based matching aims to determine the 6D pose of an object by aligning its CAD model within a 2D image of the scene, typically in colour or greyscale (e.g., Figure~\ref{fig:halcon_shape_based_matching}). The depth camera adopted for the industrial use case only outputted greyscale images.
\\[1ex]
Shape-based matching performs optimally when the object of interest appears distinct and easily discernible from its environment. While the greyscale images already provided limited contrast between the steel workpieces and their surroundings---including solid deck steel pallets, the punching machine's press bed, and grease-stained wooden pallets---dark shadows cast across the scene further exacerbated the pose estimation process. For instance, the workpieces at the bottom of the framed pallet in Figure~\ref{fig:shape_based_matching_greyscale} (top) could not be localised because the side plank's shadow created a darkened area near the products.

In Figure~\ref{fig:shape_based_matching_greyscale} (bottom), all relevant objects were successfully detected, even those at the bottom of the pallet, as indicated by the blue 3D outline of the CAD model overlaid onto its 2D representations. From the skewed stack of products in the centre, the algorithm manages to pick the top product and determine its pose. Overall, 3D shape-based matching using greyscale images demonstrated moderate robustness, handling most cases well but struggling under certain (lighting) conditions.

\begin{figure}[htb]
	\centering
	\begin{subfigure}[b]{0.48\textwidth}
		\includegraphics[width=\textwidth]{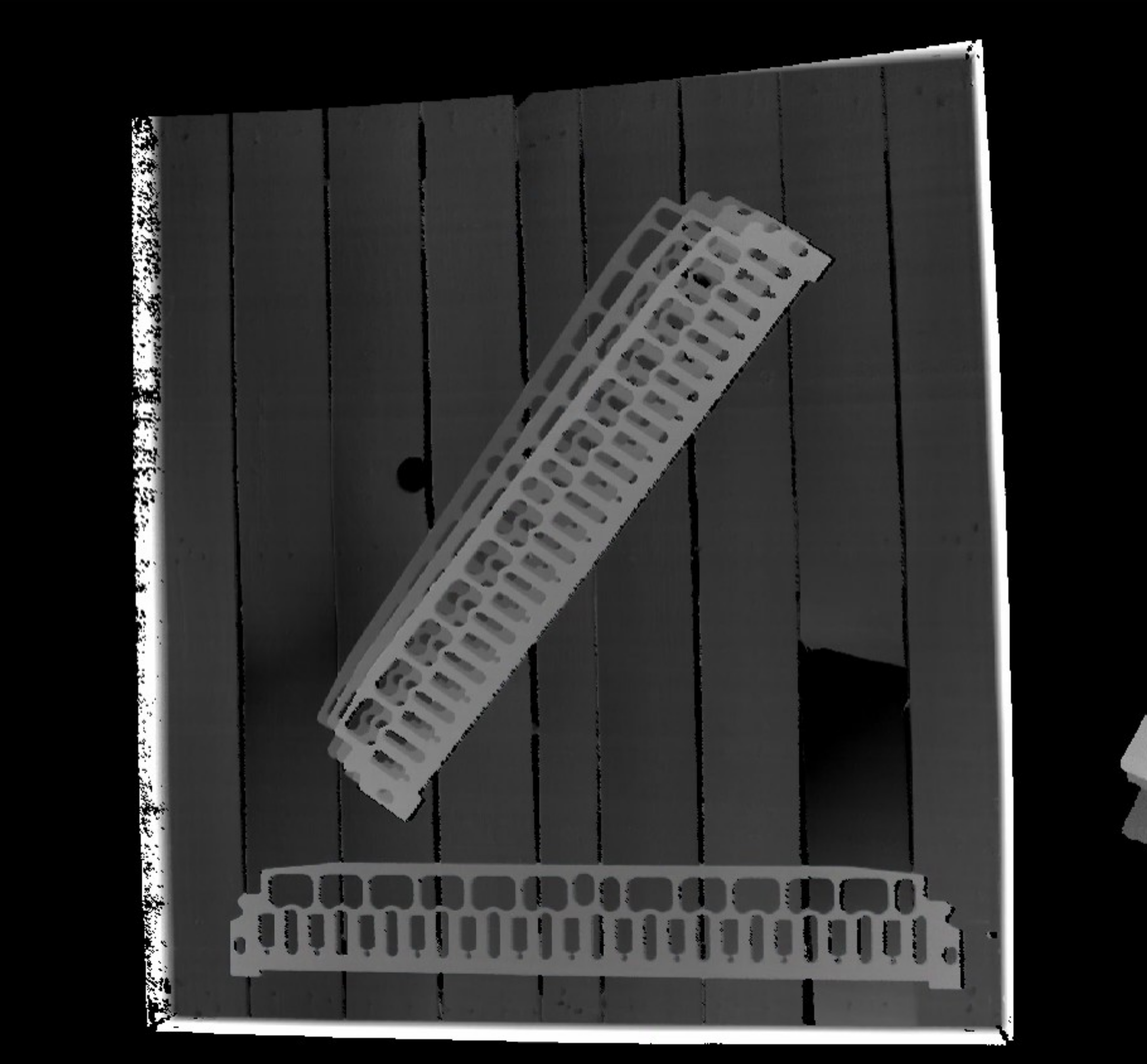}
	\end{subfigure}
	\par\smallskip
	\begin{subfigure}[b]{0.48\textwidth}
		\includegraphics[width=\textwidth]{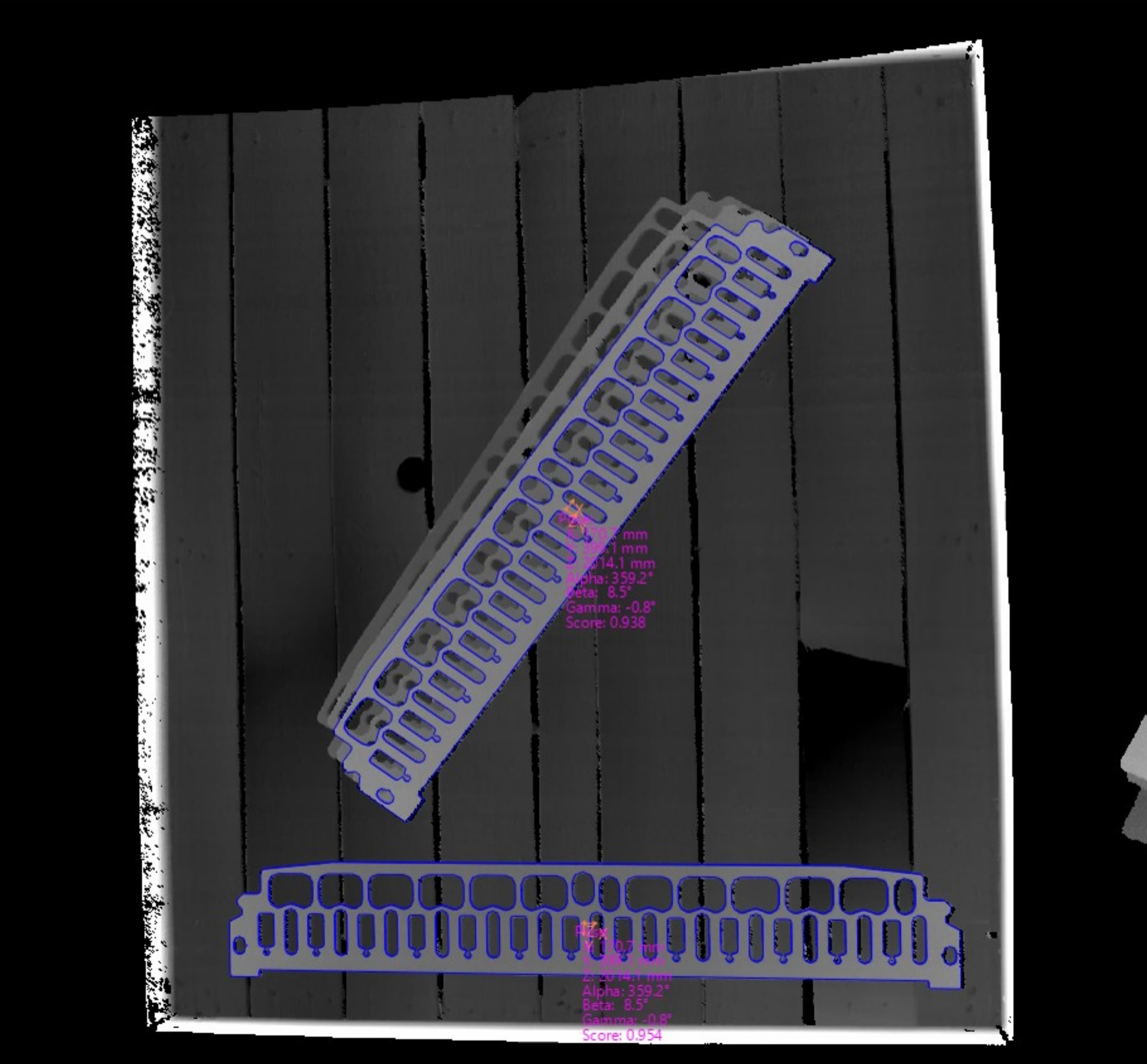}
	\end{subfigure}
	\caption{Generated depth images of framed pallets containing stacks of workpieces, with enhanced contrast (top) to improve the robustness of the 3D shape-based matching algorithm. Blue 3D outlines indicate the estimated poses of the top workpiece, closest to the camera, in each product stack (bottom).}\label{fig:shape_based_matching_depth_images}
\end{figure}

To exclude the influence of external factors---such as lighting conditions and material (dis)colouration---depth data was leveraged to enhance image contrast. More specifically, greyscale images were created in which each pixel's intensity represents the distance of its corresponding depth point relative to the camera, projected along the scene's normal vector. As such, the colour mapping in Figure~\ref{fig:shape_based_matching_depth_images} (top) ranges from white for elements belonging to the foreground (e.g., side planks) to black for the most distant ones (e.g., bottom planks), with the gradient's axis oriented perpendicular to the bottom of the pallet. If these grey tones were simply a measure of the Euclidean distance from the camera's origin, and the camera were tilted with respect to the scene, the workpieces' top surfaces would not appear as uniformly coloured as shown in Figure~\ref{fig:shape_based_matching_depth_images} but would instead exhibit a gradient too. Hence, the pixel's intensity is set to the distance of that depth point projected along the scene's normal. As a result of this strong contrast between the background and foreground, HALCON's 3D shape-based matcher successfully and reliably estimated the 6D poses of all relevant objects in each input image provided, as exemplified in Figure~\ref{fig:shape_based_matching_depth_images} (bottom).

Shape-based matching excels not only in terms of robustness but also in speed. Compared to surface-based matching, which requires computation times ranging from \num{15} to \num{20} seconds, 3D shape-based matching processes greyscale images in an average of \(2.49 \pm 0.09\) s (\(\mu \pm \text{SEM}, n = 19\)) and depth images in \(1.22 \pm 0.07\) s (\(\mu \pm \text{SEM}, n = 19\)). When dividing the processing time per image by the number of objects detected within that image, computation times\footnote{Based on the number of samples (\(n\)), it is evident that the shape-based matching algorithm failed to localise one object (\(34 \neq 35\)) in the set of greyscale images.} decrease to, respectively, \(1.29 \pm 0.13\) s (\(\mu \pm \text{SEM}, n = 34\)) and \(0.65 \pm 0.08\) s (\(\mu \pm \text{SEM}, n = 35\)). Thus, estimating the 6D pose of sheet-metal parts in the dataset of depth images takes less than two-thirds of a second. Since cycle time---i.e., the total time allocated for localising and picking a product---was amongst the most critical challenges discussed in Section~\ref{sec:challenges}, the method presented in this section was ultimately selected for deployment by the industrial client. Regarding the algorithm's own metric of accuracy, a similarly high performance was achieved, with matching scores of \(0.925 \pm 0.003\) (\(\mu \pm \text{SEM}, n = 34\)) for greyscale images and \(0.957 \pm 0.002\) \((\mu \pm \text{SEM}, n = 35\)) for depth images.

\begin{figure}
	\centering
	\includegraphics[width=\linewidth]{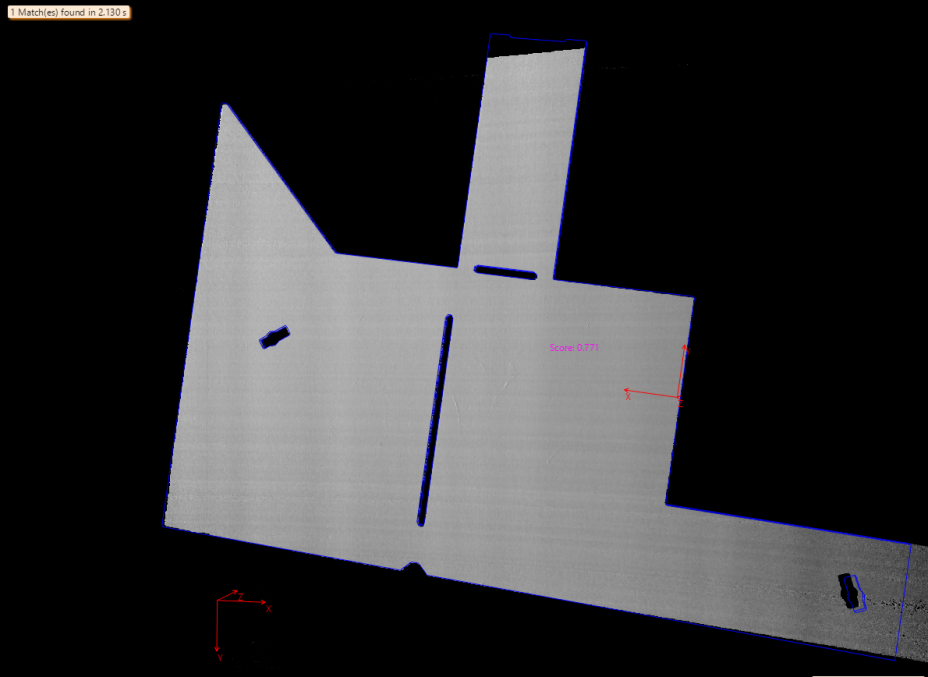}
	\caption{Generated depth image of a large product with its background removed. The blue 3D outline indicates a successful 3D shape-based matching using a truncated CAD model. Matching completed in \num{2.13} seconds with a score of \num{0.77}.}\label{fig:shape_based_matching_sizeable_product}
\end{figure}

Finally, Figure~\ref{fig:shape_based_matching_sizeable_product} illustrates that the developed technique is effective even for large products, which the depth camera could only partially capture, though by truncating the CAD model to only the visible section of the workpiece.

\section{Vision-based referencing}

\subsection{Scene-to-camera transformation}\label{sec:scene_to_camera_transformation}

As has become apparent, some of the methods addressed before could perform better or would require less pre-processing when the coordinate transformation between the camera and the scene is pre-determined. Indeed, if the pose of the framed pallet in Section~\ref{sec:z_channel_thresholding} relative to the camera would have been known, Z-channel thresholding could have been applied along the Z-axis of the pallet's coordinate frame rather than that of the depth camera. This approach would enable a more symmetrical removal of the side planks on either side of the scene. As stressed before, maximising the subtraction of background points can significantly enhance computational efficiency and improve the matching score of 3D surface-based matching. Furthermore, understanding the coordinate transformation between the depth camera and the scene can also facilitate in generating high-contrast depth images for 3D shape-based matching (see Section~\ref{sec:shape_based_matching}) by ensuring that the colour gradient's axis aligns as closely as possible with the pallet's normal vector.

\begin{figure}[htb]
	\centering
	\includegraphics[width=\linewidth]{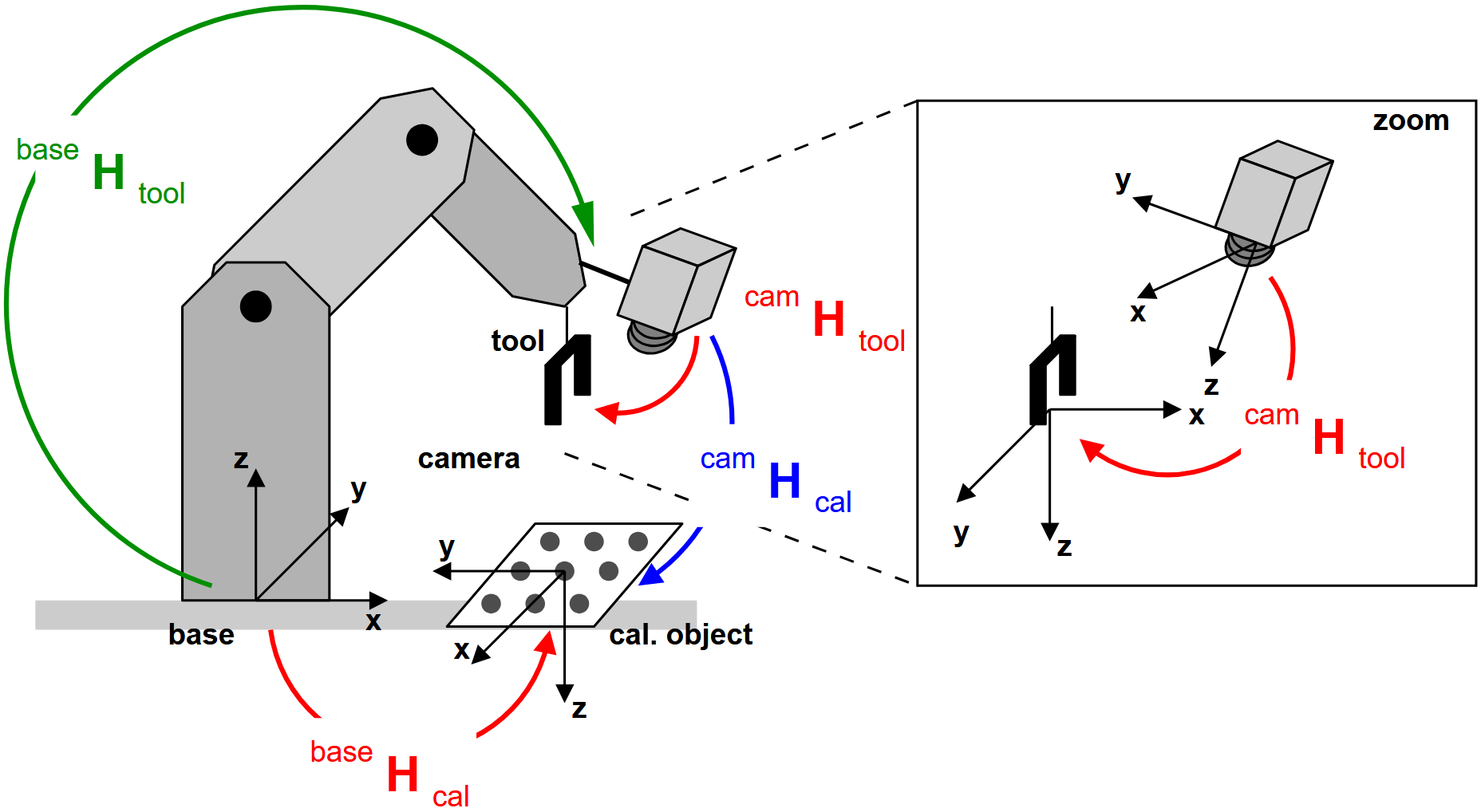}
	\caption{Kinematic chain of homogeneous transformations for calibrating a robotic \emph{eye-in-hand} system~\cite[Ch. 8]{MVTec:2025a}.}\label{fig:robot_camera_coordinate_transformations}
\end{figure}

In order to deduce the position and orientation of the scene's coordinate system with respect to the camera, a stationary calibration object could be placed within the scene, either permanently, so that the camera can capture and localise it at any time during operation, or intermittently, whenever re-calibration is required. Figure~\ref{fig:robot_camera_coordinate_transformations} illustrates such a setup with a depiction of the existing coordinate transformations between the camera, tool flange, cobot base and calibration object for an \emph{eye-in-hand} or \emph{moving-camera} configuration\footnote{Alternative arrangements may involve one or multiple static cameras or a mix of moving and fixed cameras; however, these configurations are beyond the scope of this article.}. The homogeneous transformation required for the objective outlined at the beginning of this section is denoted as \(\prescript{cam}{}H_{cal}\), representing the pose of the static calibration object---and, by extension, the scene---expressed within the camera's coordinate system.

The choice of calibration object is constrained by the type of data the depth sensor provides. When a 3D point cloud and a 2D greyscale intensity image are outputted---as is the case for the structured-light camera used before, both a 3D object and a 2D calibration plate or (ArUco) marker tag could be viable options. For instance, the static pose of a geometrically defined reference object (e.g., a 3D sphere, cube, or even the cobot's own base~\cite{Li:2024}) can be identified by executing 3D surface-based matching on the scene's 3D point cloud containing that object. The same can be accomplished with 3D shape-based matching, though by using the greyscale image as input. Due to their simplicity and robustness, one could also opt for permanently attaching a 2D calibration pattern to one of the scene's elements, such as a machine or frame. Here, HALCON's 2D matching algorithms~\cite{MVTec:2025b} or marker detectors from open-source computer vision libraries like OpenCV~\cite{OpenCV:2025} can be employed to estimate \(\prescript{cam}{}H_{cal}\).

\begin{figure}[htb]
	\centering
	\includegraphics[width=\linewidth]{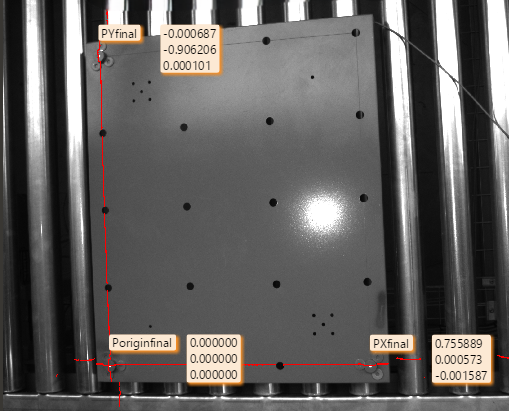}
	\caption{Greyscale image of the reference object produced by the industrial client, featuring three light beacons at its corners for establishing the roller conveyor's coordinate frame.}\label{fig:custom_calibration_rig}
\end{figure}

Finally, also custom calibration rigs may be developed in conjunction with specific alignment procedures. For the industrial application, the client designed and produced a reference plate (see Figure~\ref{fig:custom_calibration_rig}) featuring monochrome LEDs at three of its corners. To determine this calibration plate's coordinate frame, the client developed a HALCON procedure that relies on detecting the positions of these three light beacons. Their layout is pre-defined, with well-known distances between them, ensuring precise spatial relationships while remaining part of a flat plane. This method allowed for accurate referencing of the depth camera with respect to the structure on which the plate was placed, e.g., the roller conveyor.

\subsection{Object-to-cobot transformation}

Following a successful 3D object localisation, the workpiece's pose in camera coordinates must be converted to cobot base coordinates. From this, suitable control commands can be derived, enabling the cobot to perform actions such as grasping the product with a gripper. To complete this conversion, all four transformations within the cobot's kinematic chain, shown in Figure~\ref{fig:robot_camera_coordinate_transformations}, must be known. Typically, the system begins with two pre-defined homogeneous transformations:
\begin{itemize}
	\item \(\prescript{base}{}H_{tool}:\) the pose of the cobot's tool flange relative to its base, which can be acquired from the robot controller or manufacturer software;
	\item \(\prescript{cam}{}H_{cal}:\) the pose of the calibration object or plate with respect to the camera, which can be found as described in Section~\ref{sec:scene_to_camera_transformation}.
\end{itemize}

However, two coordinate transformation remain unknown before the kinematic chain can be closed. These are:
\begin{itemize}
	\item \(\prescript{cam}{}H_{tool}:\) the pose of the cobot's tool flange expressed in camera coordinates;
	\item \(\prescript{cal}{}H_{base}:\) the pose of the calibration object or plate with respect to the cobot's base frame.
\end{itemize}

For an articulated robot manipulator, hand-eye calibration provides a method to simultaneously solve for the two remaining coordinate transformations. This process involves the cobot moving the camera while capturing and localising a calibration object or plate at various poses. The resulting set of \(\prescript{cam}{}H_{cal}\) poses, combined with a corresponding set of \(\prescript{base}{}H_{tool}\) transformations from the robot controller, enable the hand-eye calibration algorithm to establish the missing links in the kinematic chain. Several toolboxes/functions are available for performing such calibration procedure, including the ones offered by HALCON~\cite[Ch. 8]{MVTec:2025a} and OpenCV~\cite{OpenCV:2025a}.

\section{Conclusion}

Due to high-mix--low-volume production, sheet-metal workshops today are challenged by small series and varying orders. As standard automation solutions tend to fall short, SMEs resort to repetitive manual labour impacting production costs and leading to tech-skilled workforces not being used to their full potential.

The COOCK+ ROBUST project aims to transform cobots into mobile and reconfigurable production assistants by integrating existing technologies, including 3D object recognition and localisation. This article has explored both the opportunities and challenges of enhancing cobotic systems with these technologies in an industrial setting, outlining the key steps involved in the process. Additionally, insights from a past project, carried out by ACRO in collaboration with an industrial partner, have served as a concrete implementation example throughout.

Stay informed about the latest developments of ROBUST and its upcoming events by following Sirris' innovation blog\footnote{https://www.sirris.be/en/inspiration/}.

\section*{Acknowledgements}

The authors sincerely appreciate the support of Flanders Innovation \& Entrepreneurship (VLAIO), whose SME e-wallet and COOCK+ subsidies made this work possible. Furthermore, they extend their gratitude to Maarten Verheyen---machine vision expert at the ACRO research unit of KU Leuven---for his valuable insights and assistance.

%------------------------------------------------
%	BIBLIOGRAPHY
%------------------------------------------------

\printbibliography[title={References}] % Print the bibliography, section title in curly brackets

@inproceedings{Drost:2010,
  author    = {Drost, Bertram and Ulrich, Markus and Navab, Nassir and Ilic, Slobodan},
  booktitle = {2010 IEEE Computer Society Conference on Computer Vision and Pattern Recognition},
  title     = {{Model globally, match locally: Efficient and robust 3D object recognition}},
  year      = {2010},
  volume    = {},
  number    = {},
  pages     = {998-1005},
  doi       = {10.1109/CVPR.2010.5540108}
}

@online{EasyODM:2024,
  title  = {{Why use YOLO object detection?}},
  author = {{EasyODM}},
  year   = {2024},
  url    = {https://easyodm.tech/yolo-object-detection/}
}

@online{JanSteel:2025,
  title  = {Sheet Metal Fabrication for Businesses Everywhere},
  author = {{JanSteel USA Inc.}},
  year   = {2025},
  url    = {https://jansteelusa.com/steel-welding-fabrication-service-company/}
}

@article{Li:2024,
  author  = {Li, Leihui and Yang, Xingyu and Wang, Riwei and Zhang, Xuping},
  year    = {2024},
  month   = {09},
  title   = {Automatic Robot Hand-Eye Calibration Enabled by Learning-Based {3D} Vision},
  journal = {Journal of Intelligent \& Robotic Systems},
  volume  = {110},
  number  = {3},
  doi     = {10.1007/s10846-024-02166-4}
}

@online{MechMind:2023,
  title  = {Brief Introduction to the Concepts of {3D} Vision System: Base Reference Frame (aka. Coordinate System)},
  author = {{Mech-Mind}},
  year   = {2023},
  url    = {https://community.mech-mind.com/t/topic/1961/}
}

@software{MVTec:2025,
  author = {{MVTec Software GmbH}},
  title  = {{HALCON - the powerful software for your machine vision application}},
  url    = {https://www.mvtec.com/products/halcon/},
  date   = {2025-05-23}
}

@manual{MVTec:2025a,
  title   = {{Solution Guide III-C: 3D Vision}},
  year    = {2025},
  author  = {{MVTec Software GmbH}},
  address = {Munich, Germany},
  url     = {https://www.mvtec.com/products/halcon/work-with-halcon/documentation/}
}

@manual{MVTec:2025b,
  title   = {{Solution Guide II-B: Matching}},
  year    = {2025},
  author  = {{MVTec Software GmbH}},
  address = {Munich, Germany},
  url     = {https://www.mvtec.com/products/halcon/work-with-halcon/documentation/}
}

@manual{OpenCV:2025,
  title  = {Detection of {ArUco} Markers},
  year   = {2025},
  author = {Garrido, Sergio and Panov, Alexander},
  url    = {https://docs.opencv.org/4.x/d5/dae/tutorial_aruco_detection.html}
}

@manual{OpenCV:2025a,
  title  = {Camera Calibration and {3D} Reconstruction},
  year   = {2025},
  author = {OpenCV},
  url    = {https://docs.opencv.org/4.x/d9/d0c/group__calib3d.html#gaebfc1c9f7434196a374c382abf43439b}
}

@online{PCL:2025,
  title  = {Removing outliers using a {StatisticalOutlierRemoval} filter},
  author = {{Point Cloud Library}},
  year   = {2025},
  url    = {https://pcl.readthedocs.io/projects/tutorials/en/latest/statistical_outlier.html}
}

@online{Photoneo:2019,
  title  = {Frequently asked questions and Frequently experienced difficulties},
  author = {{Photoneo}},
  year   = {2019},
  url    = {https://wiki.photoneo.com/index.php/Frequently_asked_questions_and_Frequently_experienced_difficulties#How_do_I_scan_reflective_materials/}
}

@online{Photoneo:2025,
  title  = {{PhoXi 3D scanner L}},
  author = {{Photoneo}},
  year   = {2025},
  url    = {https://www.photoneo.com/products/phoxi-scan-l/}
}

@online{Sirris:2025,
  title  = {Cobot on a movable platform: a solution for the sheet metal industry},
  author = {Vanongeval, Ward},
  year   = {2025},
  url    = {https://www.sirris.be/en/inspiration/advanced-manufacturing/cobot-on-movable-platform/}
}

@techreport{Sweetser:2023,
  title       = {{Optical filters for Intel® RealSense™ depth cameras D400}},
  author      = {Sweetser, John and Grunnet-Jepsen, Anders},
  year        = {2023},
  institution = {Intel RealSense}
}

@article{Wang:2022,
  author = {Wang, Chien-Yao and Bochkovskiy, Alexey and Liao, Hong-Yuan Mark},
  year   = {2022},
  month  = {07},
  pages  = {1-15},
  title  = {{YOLOv7: Trainable bag-of-freebies sets new state-of-the-art for real-time object detectors}},
  doi    = {10.48550/arXiv.2207.02696}
}

@online{WiredWorkers:2025,
  title  = {Bin picking with a collaborative robot},
  author = {{WiredWorkers}},
  year   = {2025},
  url    = {https://www.wiredworkers.io/cobot/applications/bin-picking/}
}

%------------------------------------------------

\end{document}